\documentclass[letterpaper]{article} 
\usepackage{aaai2026}  
\usepackage{times}  
\usepackage{helvet}  
\usepackage{courier}  
\usepackage[hyphens]{url}  
\usepackage{graphicx} 
\usepackage{natbib}  
\usepackage{caption} 
\usepackage[T1]{fontenc}
\usepackage{latexsym}
\usepackage{amssymb}
\usepackage{algorithm}
\usepackage{algorithmic}
\usepackage{array}
\usepackage{microtype}
\usepackage{url}
\usepackage{booktabs}
\usepackage{lineno}
\usepackage{multirow}
\usepackage{amsmath}
\usepackage{adjustbox}
\usepackage{tabularx, colortbl, xcolor}
\usepackage{subcaption}
\newcolumntype{Y}{>{\raggedright\arraybackslash}X}
\definecolor{shadecolor}{gray}{0.9}

\usepackage{newfloat}
\usepackage{listings}
\DeclareCaptionStyle{ruled}{labelfont=normalfont,labelsep=colon,strut=off} 
\floatstyle{ruled}
\newfloat{listing}{tb}{lst}{}
\floatname{listing}{Listing}
\nocopyright

\title{
  \adjustbox{valign=c}{\includegraphics[height=28pt]{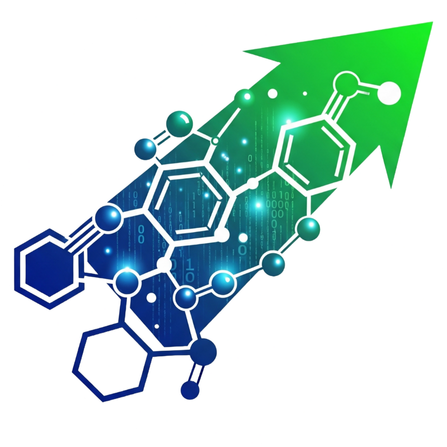}}~%
  ChemPro: A Progressive Chemistry Benchmark for Large Language Models
}

\author{
    Aaditya Baranwal,
    Shruti Vyas
}
\affiliations{
    University of Central Florida
}

\vspace{-30pt}

\makeatletter
\AtBeginDocument{
\let\@oldmaketitle\@maketitle
\renewcommand{\@maketitle}{\@oldmaketitle
  \vspace{10pt}
\begin{minipage}{0.32\textwidth}
    \includegraphics[width=\linewidth]{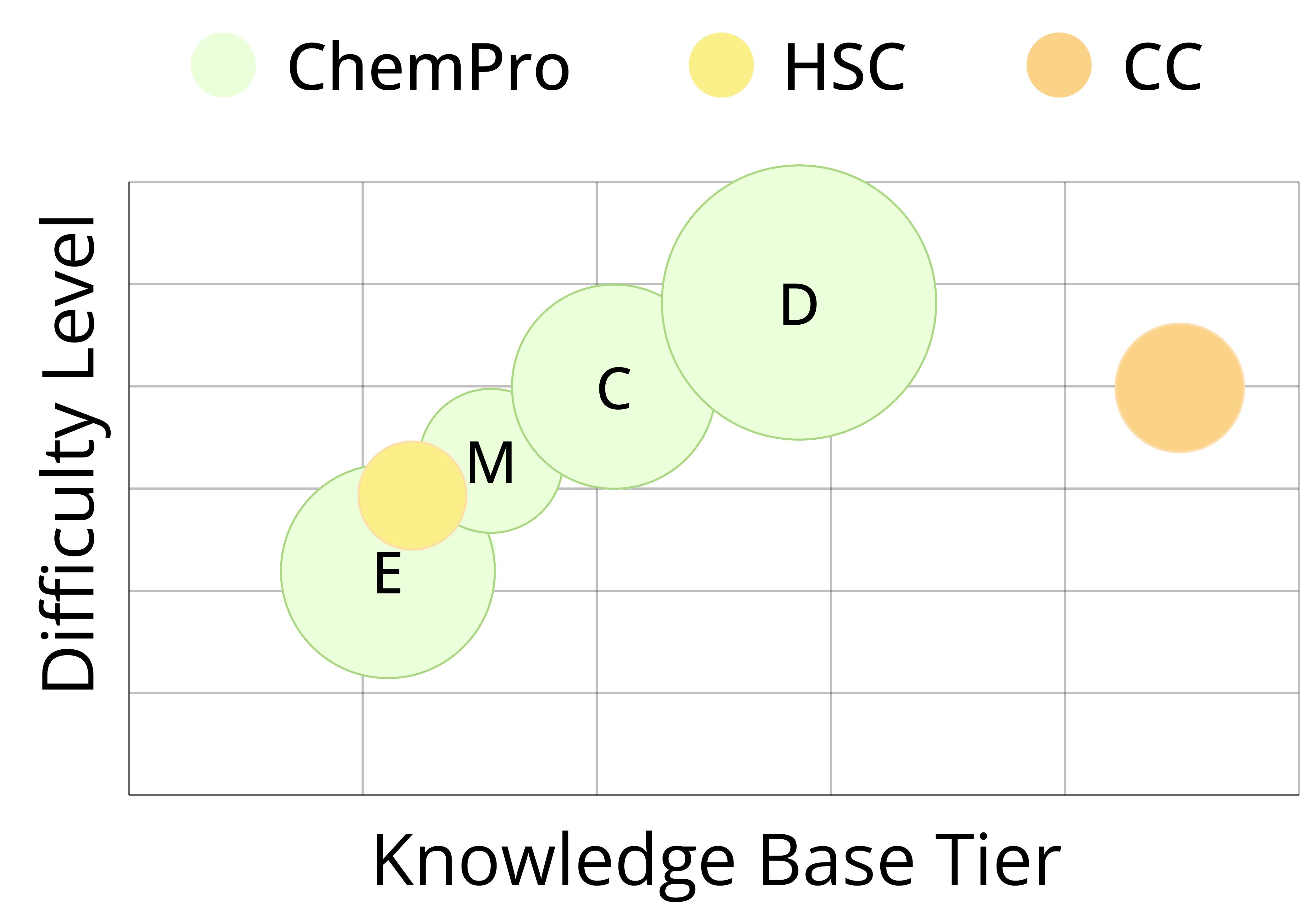}
\end{minipage} \quad
\begin{minipage}{0.33\textwidth}
    \includegraphics[width=\linewidth]{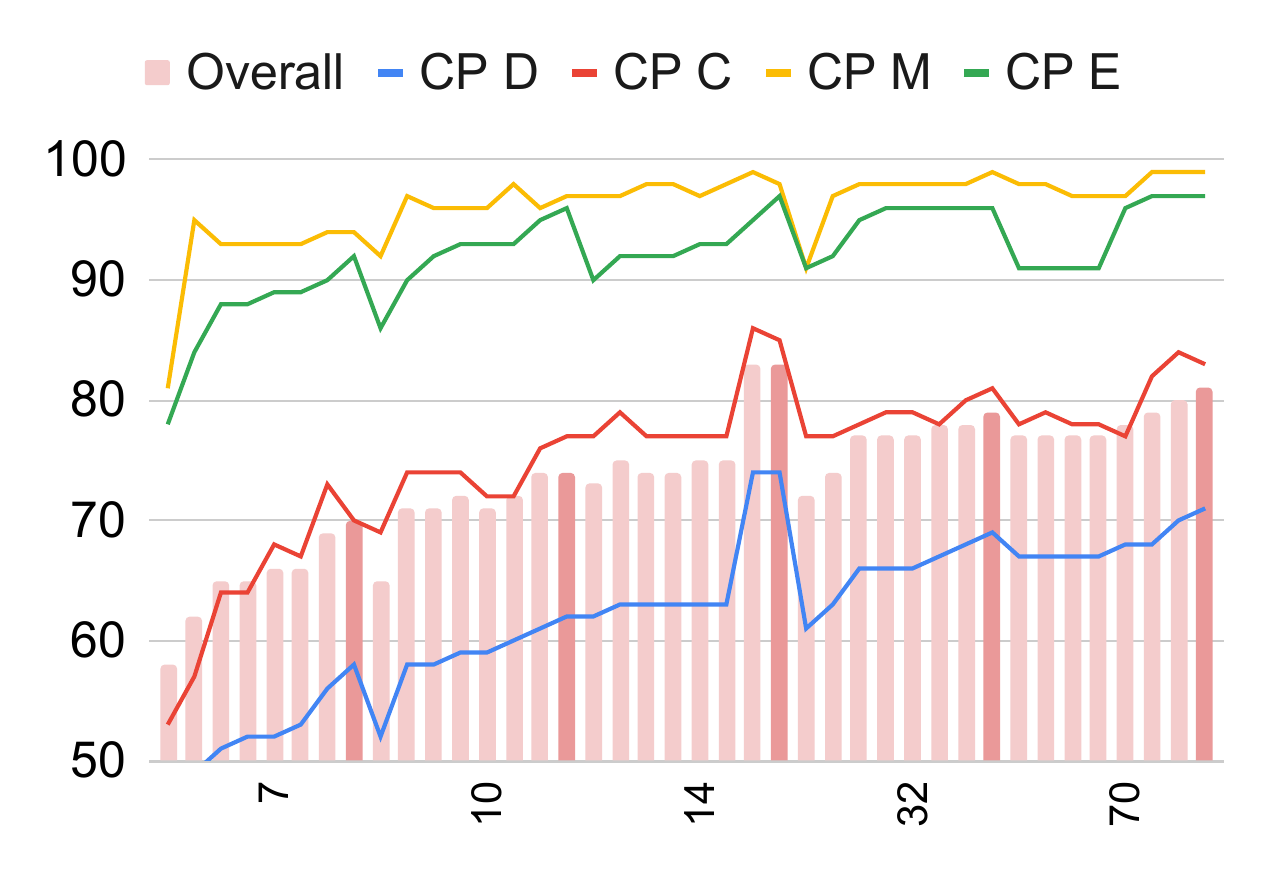}
\end{minipage} \quad
\begin{minipage}{0.32\textwidth}
    \includegraphics[width=\linewidth]{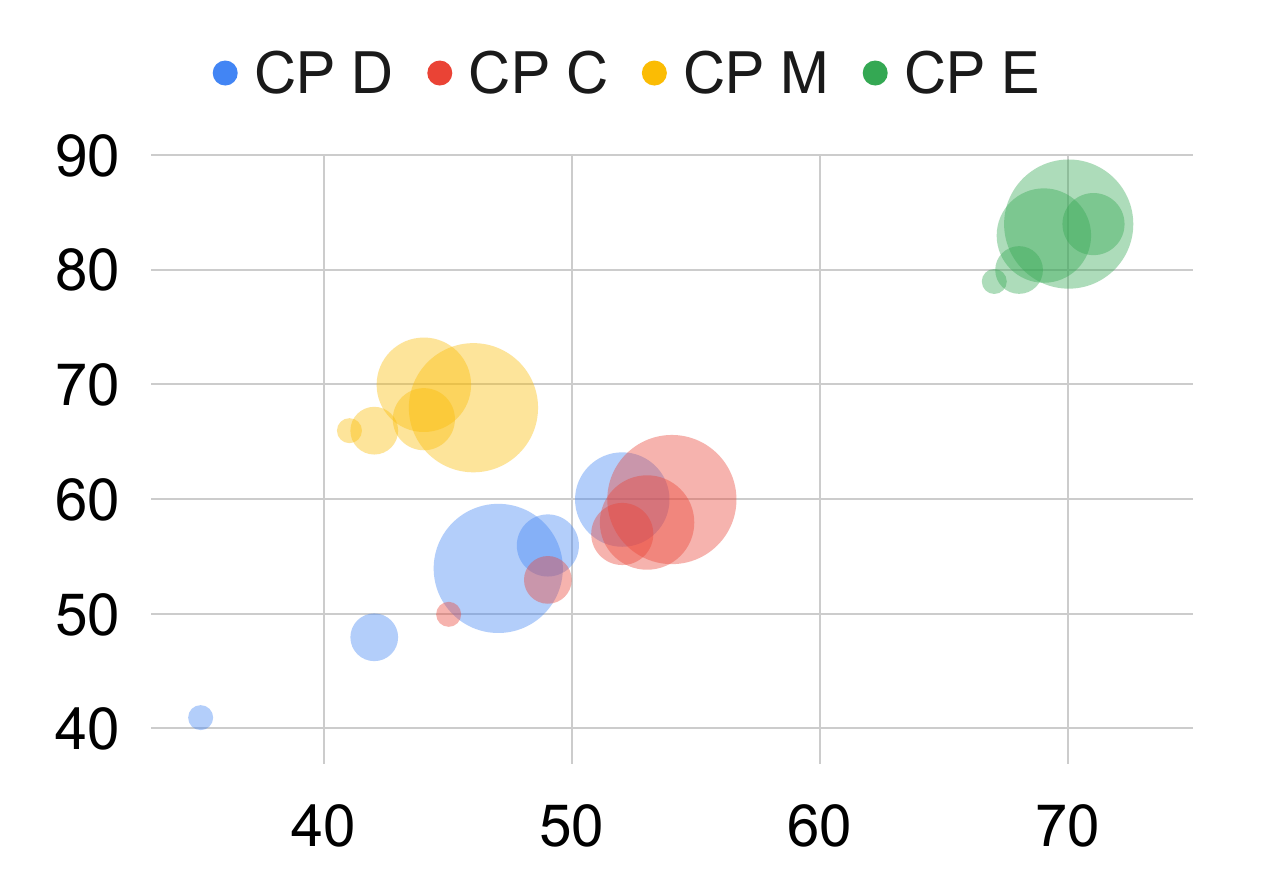}
\end{minipage}

  \vspace{5pt}
  \captionof{figure}{
    \textbf{\textit{Overview of ChemPro.}} \textbf{\textit{Left:}} A comparison with existing benchmarks. \textit{Y-axis} represents LLM difficulty (lower mean MCQ accuracy across models = harder). \textit{X-axis} represents academic succession (Elementary to Graduate/Expert), rendered as a continuum because real-world curricula overlap across grade boundaries. \textit{Bubble size} represents question count. \textbf{E}, \textbf{M}, \textbf{C}, and \textbf{D} are ChemPro's \textbf{E}asy, \textbf{M}edium, \textbf{C}hallenging, and \textbf{D}ifficult sections (axis derivation details in supplementary).
    \textbf{\textit{Center:}} Performance (Accuracy, y-axis) of all 40 open-source models evaluated on ChemPro MCQs showing the impact of model-size (x-axis)  on performance (lines are Performance on individual ChemPro sections and columns is the overall average performance).
    \textbf{\textit{Right:}} Exact-match accuracy (x-axis) vs Tolerance-based accuracy (y-axis) on ChemPro Numerical for all 40 open-source models (bubble size represents model parameter count).
  \label{fig:teaser}
  \vspace{10pt}
 }
}}

\begin{document}

 \maketitle
\begin{abstract}
We introduce \textbf{\textit{ChemPro}}, a progressive benchmark with 4100 natural language question-answer pairs in Chemistry, across 4 coherent sections of \textit{difficulty} designed to assess the proficiency of Large Language Models (LLMs) in a broad spectrum of general chemistry topics. We include Multiple Choice Questions and Numerical Questions spread across fine-grained information recall, long-horizon reasoning, multi-concept questions, problem-solving with nuanced articulation, and straightforward questions in a balanced ratio, effectively covering \textit{Bio-Chemistry}, \textit{Inorganic-Chemistry}, \textit{Organic-Chemistry} and \textit{Physical-Chemistry}. \textbf{\textit{ChemPro}} is carefully designed analogous to a student's academic evaluation for basic to high-school chemistry. A gradual increase in the question \textit{difficulty} rigorously tests the ability of LLMs to progress from solving basic problems to solving more sophisticated challenges.

We evaluate 45+7 state-of-the-art LLMs, spanning both open-source and proprietary variants, and our analysis reveals that while LLMs perform well on basic chemistry questions, their accuracy declines with different types and levels of complexity. These findings highlight the critical limitations of LLMs in general scientific reasoning and understanding and point towards understudied dimensions of difficulty, emphasizing the need for more robust methodologies to improve LLMs.
\end{abstract}    
\section{Introduction}

\begin{table*}[t]
\small
\caption{\textbf{\textit{Comparison with chemistry and related science benchmarks.}} ChemPro uniquely provides \textit{source-aligned difficulty provenance} and comprehensive \textit{elementary-to-high-school} coverage with dual assessment modes. \textit{Knowledge Tier}: Ex=Expert/Research, UG=Undergraduate, HS=High School, E=Elementary.}
\label{tab:benchmark_comparison}
\vspace{-5pt}
\begin{center}
\begin{tabular}{|l|l|l|l|c|}
\toprule
\textbf{Benchmark} & \textbf{Topics} & \textbf{Format} & \textbf{Knowledge Tier} & \textbf{Progressive} \\
\midrule
ARC ~\citep{clark2018thinkgothought} & Science & MCQs & E-HS & No \\
ScienceQA ~\citep{lu2022learnexplainmultimodalreasoning} & Science & MCQs & E-HS & No \\
MMLU ~\citep{hendrycks2021measuringmassivemultitasklanguage} & Multi-domain & MCQs & UG-Ex & No \\
GPQA ~\citep{rein2023gpqagraduatelevelgoogleproofqa} & Multi-domain & MCQs & Graduate-Ex & No \\
JEEBench ~\citep{arora2023llmsadvancedenoughchallenging} & STEM & MCQ+Open & HS-Advanced & No \\
SciBench ~\citep{wang2023scibenchcollegelevelscientific} & STEM & MCQ+Open & UG-Advanced & No \\
\midrule
SMolInstruct ~\citep{yu2024smolinstruct} & Chemistry & Instruction & UG-Research & No \\
MolInstruct ~\citep{ye2024molinstruct} & Chemistry & Instruction & UG-Research & No \\
CACTUS ~\citep{mcnaughton2024cactuschemistryagentbenchmark} & Chemistry & Agent Tasks & UG-Research & No \\
ScholarChemQA~\citep{chen2024scholarchemqaunveilingpowerlanguage} & Chemistry & Literature & Advanced-Research & No \\
ChemBench ~\citep{mirza2024largelanguagemodelssuperhuman} & Chemistry & MCQ+Open & UG-Graduate & No \\
ChemLLMBench ~\citep{guo2023largelanguagemodelschemistry} & Chemistry & Open+Code & Advanced-Research & No \\
RESTEEM ~\citep{song2024resteemdatarepositoryeducational} & Chemistry & Educational & E-UG & No \\
HS Chemistry ~\citep{taylor2022galacticalargelanguagemodel} & Chemistry & MCQs & HS & No \\
College Chemistry ~\citep{taylor2022galacticalargelanguagemodel} & Chemistry & MCQs & UG-Graduate & No \\
\midrule
\textbf{ChemPro (Ours)} & \textbf{Chemistry} & \textbf{MCQ + Numerical} & \textbf{E-HS} & \textbf{Yes} \\
\bottomrule
\end{tabular}
\end{center}
\vspace{-10pt}
\end{table*}

LLMs have demonstrated strong capabilities in language, coding, mathematics, and physics \citep{minaee2024largelanguagemodelssurvey,jiang2024surveylargelanguagemodels,ahn2024largelanguagemodelsmathematical,zhang2024comprehensive}. However, robust scientific reasoning remains under-evaluated at the foundational level, where conceptual understanding, numerical reasoning, and multi-step problem solving are required.

Existing  LLMs fail to adequately meet the requirements of significantly assisting professional scientific research. The LLMs are not high-agency in the context of emergent properties and abilities for science. Often called the \textit{central science} \citep{brown2017chemistry}, chemistry requires linguistic comprehension alongside symbolic manipulation (e.g., equations), multi-step calculations, and reasoning over reaction mechanisms, and is not bound under a single formal language \citep{12,guo2023chemistry}. Despite its importance, chemistry has received comparatively less benchmark attention \citep{11,guo2023chemistry}, leaving a gap in evaluating foundational proficiency beyond rule-bound domains like programming and mathematics \citep{1,4,5}.

\begin{figure*}[t!]
  \centering  
  \includegraphics[width=\linewidth]{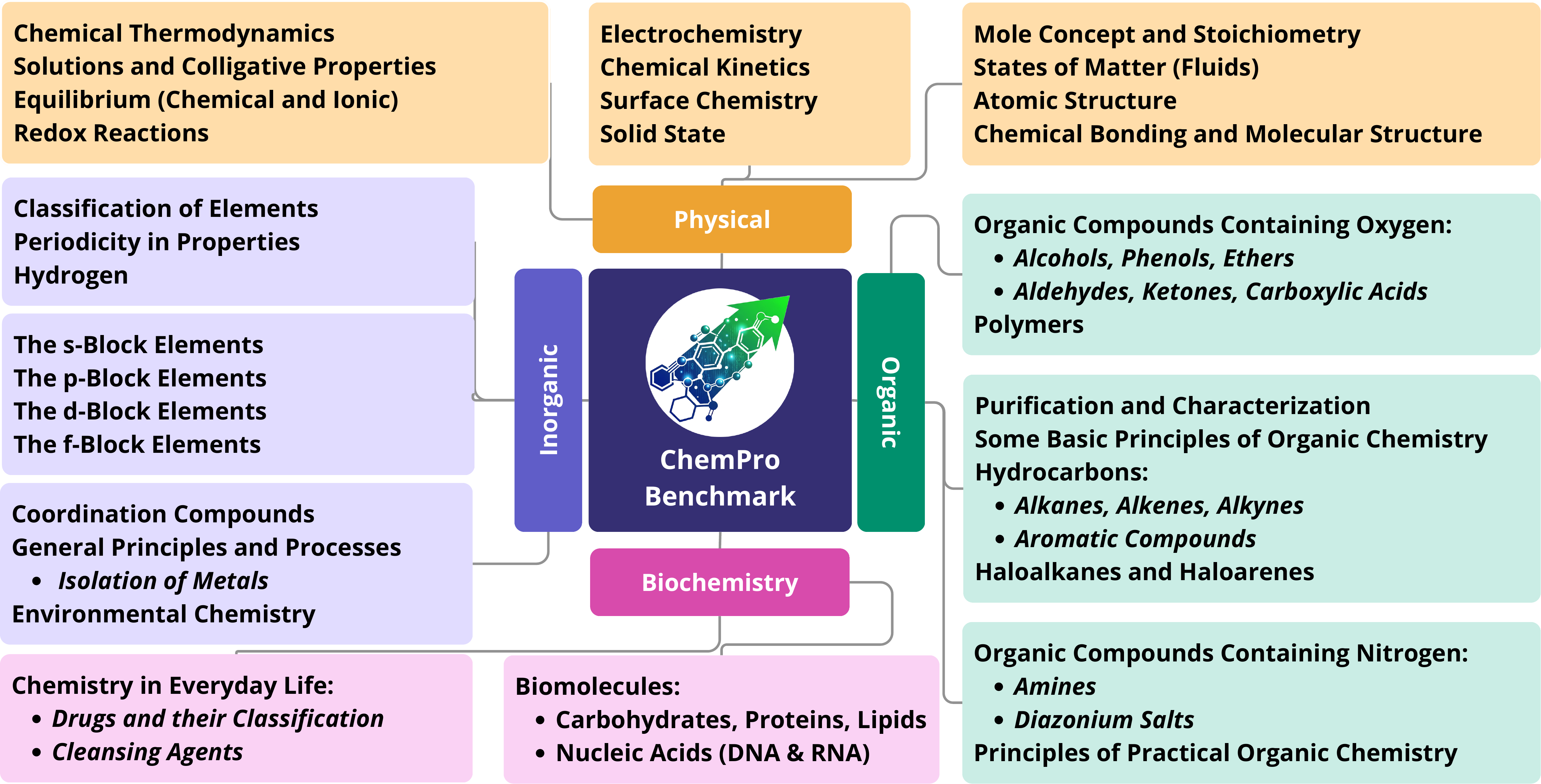}   
  \vspace{-15pt}
  \caption{\textbf{\textit{ChemPro benchmark structure:}} The benchmark spans four chemistry subfields (Biochemistry, Inorganic, Organic, Physical Chemistry) across four sections of difficulty ($\mathcal{CP}_E$, $\mathcal{CP}_M$, $\mathcal{CP}_C$, $\mathcal{CP}_D$) with balanced distribution of MCQs and numerical problems. Complete category distribution details are provided in Appendix.}
  \label{fig:chempro_subfields}
  \vspace{-5pt}
\end{figure*}

To bridge this gap, we introduce \textbf{\textit{ChemPro}} (Figure \ref{fig:chempro_subfields}), a progressive benchmark designed to evaluate LLM chemistry proficiency via a curriculum-aligned progression of questions \citep{13}. ChemPro\footnote{We use $\mathcal{CP}$ as acronym for ChemPro everywhere in the paper.} consists of 4,100 questions sourced from standardized materials including competitive exams \citep{jee_official}, textbooks \citep{ncert_official}, and quizlets \citep{edu_quizzes}, spanning Inorganic, Organic, Physical and Bio-chemistry.

We conduct a comprehensive evaluation of state-of-the-art LLMs \citep{sota1,sota2,phi4,sota4,adak2025molvision} on ChemPro, analyzing their performance across a diverse set of chemistry topics (Figure \ref{fig:teaser}). Our experiments span 45 different models, exploring proprietary and open-source variants with varying model sizes.

\textbf{Coverage and provenance.} ChemPro uses educational provenance (NCERT and JEE Mains) to enforce \emph{difficulty ordering} while spanning high-school chemistry; results are reported per tier and subfield.

\vspace{-0.5em}
\section{Related Work}

\noindent \textbf{General-Purpose and STEM Benchmarks:}
Foundation models have led to specialized benchmarks across domains. In mathematics, MATH \citep{hendrycks2020measuringmathematicalproblemsolving} evaluates symbolic reasoning, while programming benchmarks like HumanEval \citep{chen2021evaluatinglargelanguagemodels} assess code generation. Multi-domain benchmarks include MMLU \citep{hendrycks2021measuringmassivemultitasklanguage} and GPQA \citep{rein2023gpqagraduatelevelgoogleproofqa} targeting graduate-level questions. For scientific reasoning, ARC \citep{clark2018thinkgothought} provides elementary to high-school science questions but lacks chemistry-specific depth. SciBench \citep{wang2023scibenchcollegelevelscientific} evaluates college-level scientific problem-solving but focuses on undergraduate-to-graduate content. JEEBench \citep{arora2023llmsadvancedenoughchallenging} evaluates high-school problems but lacks systematic difficulty progression within subjects.

\noindent
\textbf{Chemistry-Specific Benchmarks:}
Recent chemistry-focused models like ChemLactica \citep{castro2024chemlatica} and Llama-Chem \citep{feng2024llamachemllmlargelanguage} demonstrate domain expertise but lack systematic evaluation across curriculum-aligned difficulty levels. Molecular-focused benchmarks include SMolInstruct \citep{yu2024smolinstruct}, MolInstruct \citep{ye2024molinstruct}, and CACTUS \citep{mcnaughton2024cactuschemistryagentbenchmark}, which target expert-level capabilities rather than fundamentals. Research Literature focused benchmarks include ScholarChemQA \citep{chen2024scholarchemqaunveilingpowerlanguage} and ChemBench \citep{mirza2024largelanguagemodelssuperhuman}, are focued on advanced topics and lack structure for generalisability and systematic difficulty assessment.
\noindent \textbf{Critical Limitations:}
Current chemistry evaluation suffers from three fundamental gaps: (1) \textbf{Difficulty annotation subjectivity}-most benchmarks use broad categorizations lacking verifiable educational grounding; (2) \textbf{Inadequate foundational coverage}-existing benchmarks target specialized tasks, neglecting systematic elementary-to-high-school evaluation; (3) \textbf{Limited assessment modalities}-few benchmarks combine conceptual understanding (MCQs) with computational reasoning (numerical problems).
Structured comparison between relevant (STEM and chemistry) benchmarks and ChemPro is provided in Table \ref{tab:benchmark_comparison}.

\noindent \textbf{ChemPro's Positioning:}
ChemPro addresses these limitations through: \textbf{Source-aligned difficulty provenance} tied to established curricula (NCERT) and examinations (JEE), providing verifiable educational ordering; \textbf{Comprehensive E-HS focus} systematically evaluating foundational chemistry within educational boundaries; \textbf{Dual-mode assessment} with multiple choice questions and numerical problems for comprehensive evaluation of LLMs.

\section{ChemPro Benchmark}

\begin{table*}[t!]
    \centering
    \small
    \caption{\textbf{\textit{Performance across ChemPro MCQ difficulty levels:}} Top 3 performing models across ChemPro MCQ sections by model size category. Performance metrics show accuracy scores for each section with progressive difficulty from Easy to Difficult. Note the systematic performance degradation (highlighted in red) as difficulty increases. (\textit{P: Proprietary})}
    \vspace{-5pt}
    \resizebox{0.8\textwidth}{!}{
    \begin{tabular}{|l|c|c|c|c|c|}
    \toprule
    \textbf{Best Models} & \textbf{Size} & \textbf{$\mathcal{CP}_E$} & \textbf{$\mathcal{CP}_M$} & \textbf{$\mathcal{CP}_C$ $\downarrow$} & \textbf{$\mathcal{CP}_D$ $\downdownarrows$} \\
    \midrule
    HomerCreativeAnvita-Mix-Qw7B ~\citep{suayptalha2025homercreativeanvita}                  & \multirow{3}{*}{7B} & 0.89 & 0.93 & 0.67 & \textcolor{red}{0.53} \\
    Falcon3-Jessi-v0.4-7B-Slerp  ~\citep{suayptalha2025falcon3jessi}                     & & 0.90 & 0.94 & 0.73 & \textcolor{red}{0.56} \\
    Falcon3 7B Instruct ~\citep{tiiuae2025falcon37binstruct}                        & & 0.92 & 0.94 & 0.70 & \textcolor{red}{0.58} \\
    \midrule
    MT-Merge4-gemma-2-9B        ~\citep{zelk122025mtmerge4}                    & \multirow{3}{*}{10B} & 0.93 & 0.98 & 0.72 & \textcolor{red}{0.60} \\
    falcon3-10b-tensopolis-v1   ~\citep{tensopolis2025falcon3}                     & & 0.95 & 0.96 & 0.76 & \textcolor{red}{0.61} \\
    Falcon3 10B Instruct ~\citep{tiiuae2025falcon310binstruct}                       & & 0.96 & 0.97 & 0.77 & \textcolor{red}{0.62} \\
    \midrule
    Lamarckvergence-14B       ~\citep{suayptalha2025lamarckvergence14b}                       & \multirow{3}{*}{14B} & 0.93 & 0.98 & 0.77 & \textcolor{red}{0.63} \\
    Phi-4-Model-Stock-v4      ~\citep{bunnycore2025phi4modelstockv4}                       & & 0.95 & 0.99 & 0.80 & \textcolor{red}{0.67} \\
    Luminis-PHI-4 ~\citep{suayptalha2025luminisphi4}                               & & 0.97 & 0.98 & 0.85 & \textcolor{red}{0.74} \\
    \midrule
    PathFinderAi3.0            ~\citep{daemontatox2025pathfinderai30}                      & \multirow{3}{*}{32B} & 0.96 & 0.98 & 0.78 & \textcolor{red}{0.67} \\
    Qwen2.5-32B-Instruct-abliterated-v2   ~\citep{zetasepic2025qwen2532binstructabliteratedv2}           & & 0.96 & 0.98 & 0.80 & \textcolor{red}{0.68} \\
    Rombos-LLM-V2.5-Qwen-32B ~\citep{Rombo}                   & & 0.96 & 0.99 & 0.81 & \textcolor{red}{0.69} \\
    \midrule
    shuttle-3        ~\citep{shuttleai2025shuttle3}                                & \multirow{3}{*}{70B} & 0.97 & 0.99 & 0.82 & \textcolor{red}{0.68} \\
    Homer-v1.0-Qwen2.5-72B      ~\citep{newsbang2025homerv10qwen2572b}                     & & 0.97 & 0.99 & 0.84 & \textcolor{red}{0.70} \\
    Rombos-LLM-V2.5-Qwen-72B ~\citep{Rombo}                   & & 0.97 & 0.99 & 0.83 & \textcolor{red}{0.71} \\
    \midrule
    OpenAI o1-mini   ~\citep{openai2024openaio1card}            & \multirow{3}{*}{P}             & 0.97 & 0.98 & 0.83 & \textcolor{red}{0.75} \\
    OpenAI o3-mini  ~\citep{openai2025o3mini}                               & & 0.96 & 0.99 & 0.84 & \textcolor{red}{0.75} \\
    OpenAI o1 ~\citep{openai2024openaio1card}               & & 0.97 & 0.99 & 0.85 & \textcolor{red}{0.76} \\
    \bottomrule
    \end{tabular}
    }
    \label{tab:model_performance_mcq}
\end{table*}

ChemPro is a curriculum-aligned progressive benchmark designed to systematically evaluate LLM chemistry proficiency across elementary-to-high-school difficulty levels.

\subsection{Benchmark Definition and Curation Framework}

\textbf{Formulation of ChemPro.} Let $\mathcal{L}: \mathcal{Q} \rightarrow \mathcal{A}$ represent an LLM function mapping questions to answers. We define ChemPro through a systematic curation process:

\noindent We define source spaces $\mathcal{S} = \{\mathcal{S}_E, \mathcal{S}_M, \mathcal{S}_C, \mathcal{S}_D\}$ where:
\begin{align}
\mathcal{S}_E &= \{s \in \text{Web} : \text{difficulty}(s) = \text{Elementary}\} \\
\mathcal{S}_M &= \{s \in \text{NCERT}_{9-10} : \text{grade}(s) \in [9,10]\} \\
\mathcal{S}_C &= \{s \in \text{NCERT}_{11-12} : \text{grade}(s) \in [11,12]\} \\
\mathcal{S}_D &= \{s \in \text{JEE}_{2020-2024} : \text{year}(s) \in [2020,2024]\}
\end{align}

\noindent For each question $q \in \mathcal{Q}$, validation $\mathcal{V}: \mathcal{Q} \rightarrow \{0,1\}$:
\begin{equation}
\mathcal{V}(q) = \mathcal{V}_{\text{source}}(q) \wedge \mathcal{V}_{\text{expert}}(q) \wedge \mathcal{V}_{\text{AI}}(q)
\end{equation}
where $\mathcal{V}_{\text{source}}$, $\mathcal{V}_{\text{expert}}$, and $\mathcal{V}_{\text{AI}}$ represent source verification, expert review, and AI-assisted validation respectively.

\textbf{Verification stages.} $\mathcal{V}_{\text{source}}$ enforces provenance/traceability, $\mathcal{V}_{\text{expert}}$ enforces correctness and unambiguity after textual adaptation, and $\mathcal{V}_{\text{AI}}$ performs automated consistency checks (formatting, deduplication, leakage flags); full criteria are in the supplementary.

\noindent  The final benchmark construction follows:
\begin{equation}
\mathcal{CP} = \bigcup_{i \in \{E,M,C,D\}} \mathcal{CP}_i \text{ where } \mathcal{CP}_i = \{q \in \mathcal{S}_i : \mathcal{V}(q) = 1\}
\end{equation}

\textbf{Assessment Structure:} Each $\mathcal{CP}_i$ is partitioned into:
\begin{equation}
\mathcal{CP}_i = \mathcal{M}_i \cup \mathcal{N}_i \text{ with } \mathcal{M}_i \cap \mathcal{N}_i = \emptyset
\end{equation}
where $\mathcal{M}_i$ denotes multiple-choice questions and $\mathcal{N}_i$ denotes numerical problems, ensuring coverage of both conceptual understanding and computational reasoning.

The alignment with educational sources  ensures:
\begin{equation}
\mathcal{CP}_E \prec \mathcal{CP}_M \prec \mathcal{CP}_C \prec \mathcal{CP}_D
\end{equation}
where $\prec$ denotes curriculum-verified difficulty progression.

\textbf{Source-Aligned Difficulty Provenance.} Unlike existing benchmarks with loosely defined difficulty levels, our sections are intrinsically tied to established educational sources that represent statistical consensus across thousands of educators and years of curriculum refinement:
$\mathcal{CP}_E$: Web sources, online quizlets, questionnaires, covering elementary concepts.
$\mathcal{CP}_M$: intermediate understanding.
$\mathcal{CP}_C$: advanced high-school level.
$\mathcal{CP}_D$: competitive-level problem solving.

\textbf{Curriculum Consistency Validation.} JEE Mains examination follows the official NCERT syllabus as mandated by the National Testing Agency, ensuring that $\mathcal{CP}_C$ and $\mathcal{CP}_D$ questions assess identical conceptual boundaries. The systematic performance differences between these sections (average 13-point accuracy drop from $\mathcal{CP}_C$ to $\mathcal{CP}_D$ across all models) therefore reflect variations in question formulation complexity rather than conceptual scope expansion.

\textbf{Articulation complexity.} We use this term to denote formulation-induced complexity (e.g., chained reasoning steps, conversions, cross-condition integration, and multi-concept coupling) beyond the concept label; tiering acts as a provenance-based proxy (details in supplementary).

\textbf{Statistical Reliability Over Annotator Judgments.} This source-aligned approach provides complexity measures that are orders of magnitude more reliable than individual annotator ratings, eliminating subjectivity inherent in expert and now-popular AI annotations while ensuring difficulty progression reflects genuine educational complexity.

\begin{figure}[ht]
\centering
\includegraphics[width=0.8\linewidth]{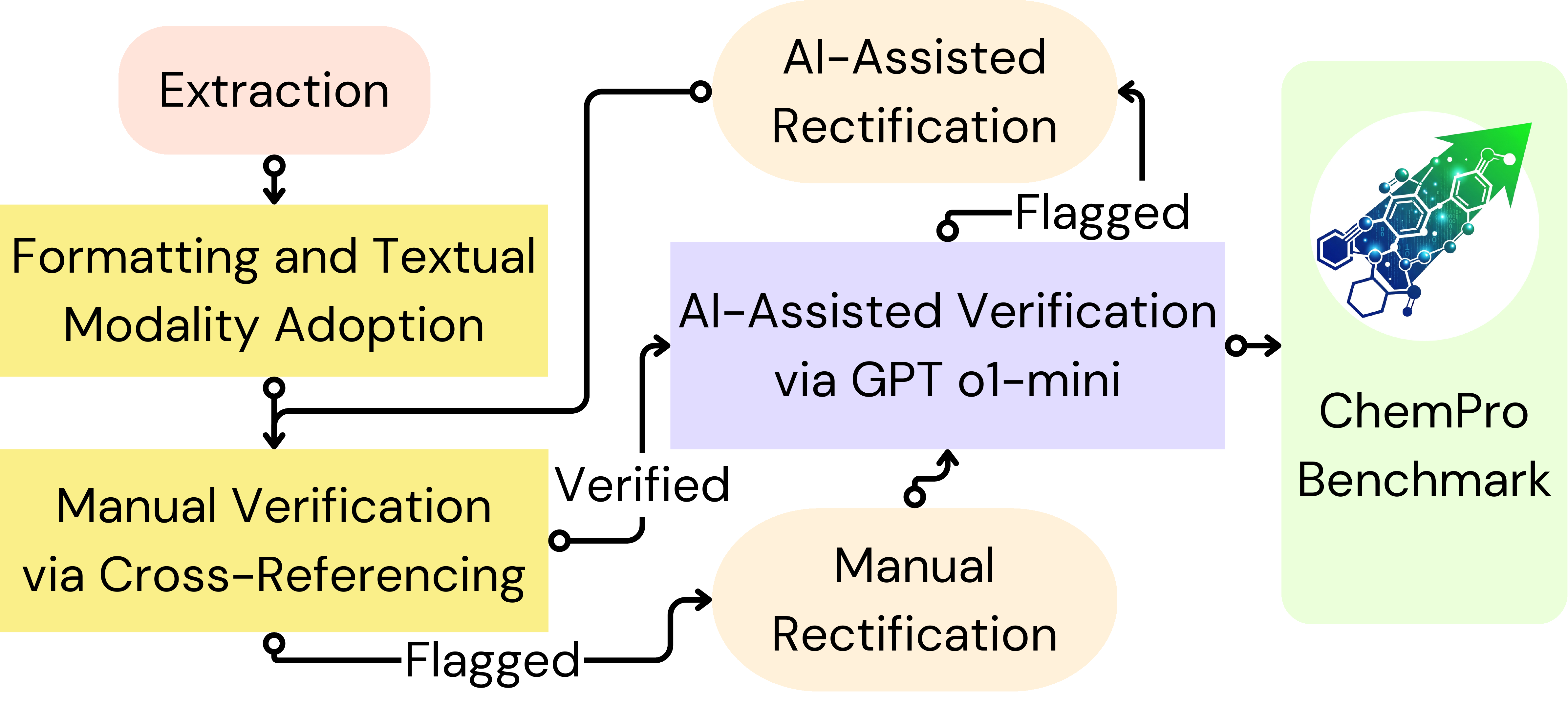}
\caption{\textbf{\textit{ChemPro Benchmark Curation Process.}} Visual workflow showing the systematic approach for creating ChemPro benchmark, from source collection across different difficulty tiers to quality validation and final dataset compilation. The process ensures source-aligned difficulty provenance while maintaining rigorous quality standards through multiple validation layers (both AI and Human).}
\vspace{-1em}
\label{fig:curation_process}
\end{figure}

\textbf{Performance Validation.} The systematic 13-point accuracy drop between $\mathcal{CP}_C$ and $\mathcal{CP}_D$ across all 45 evaluated models, despite curriculum equivalence, provides empirical evidence that observed difficulty stems from formulation complexity rather than conceptual scope differences.

\begin{figure*}[t!]
    \includegraphics[width=0.33\linewidth]{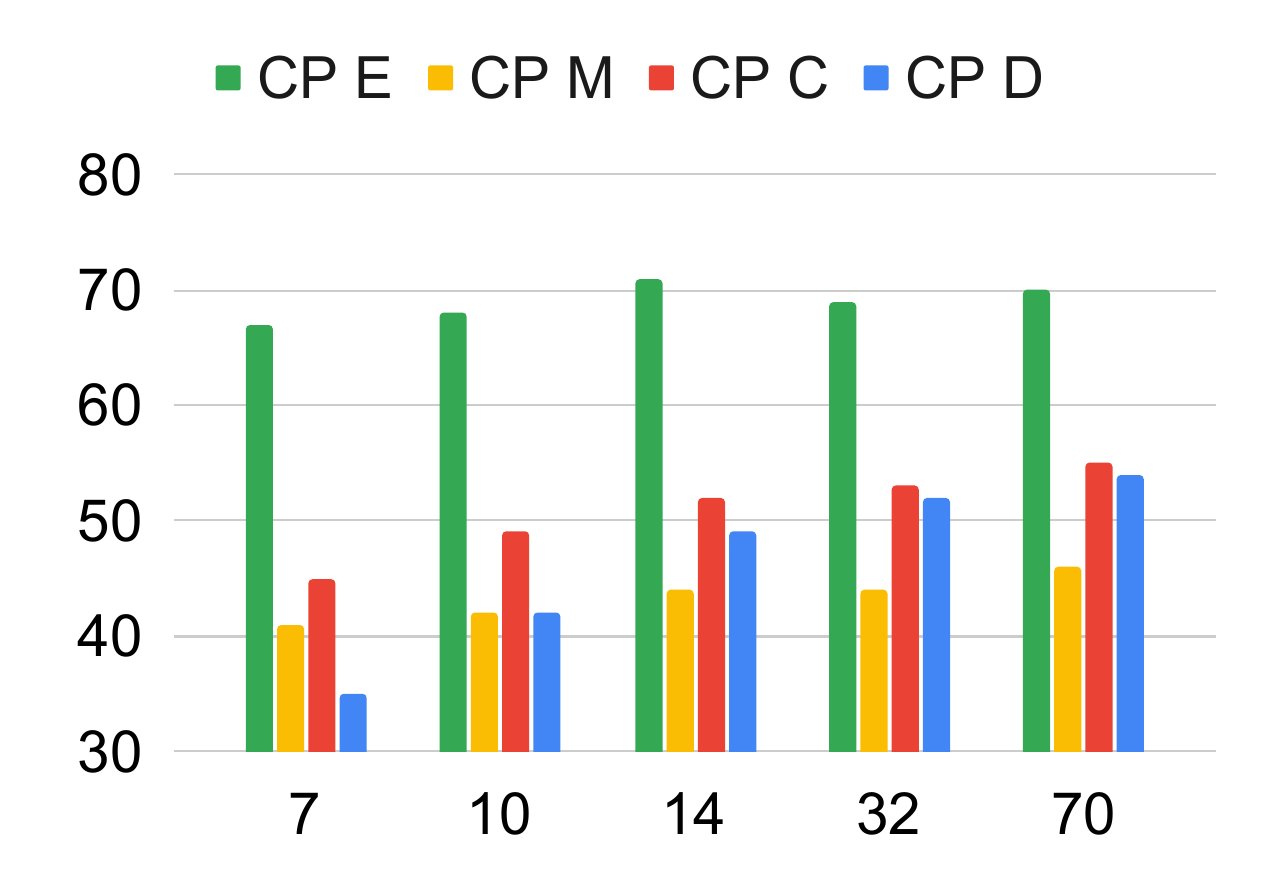}
    \includegraphics[width=0.33\linewidth]{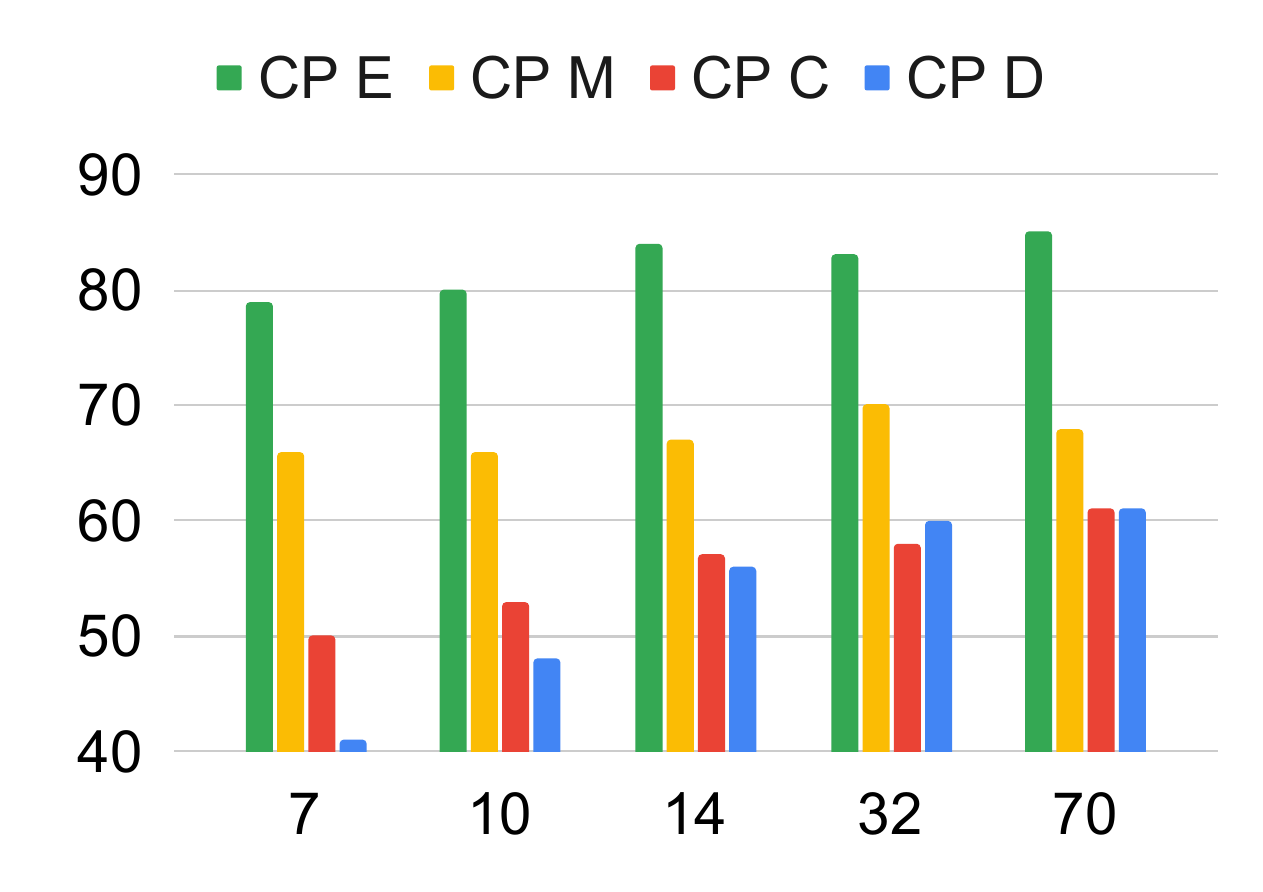}
    \includegraphics[width=0.33\linewidth]{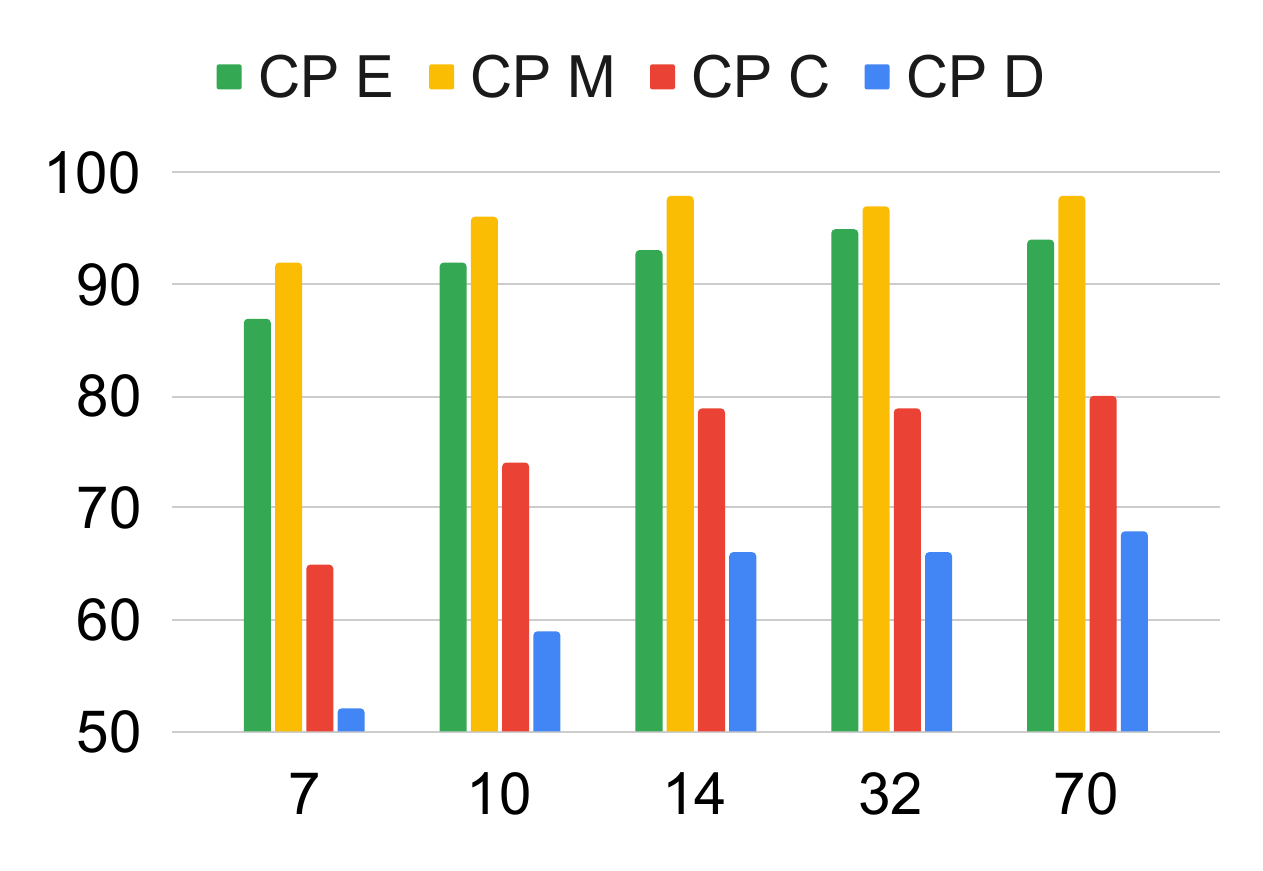}
    \caption{
    \textbf{\textit{Parameter Scaling Limitations.}}
    Scaling benefits plateau well below human performance levels (90\%+ expected accuracy).
    Numerical reasoning limitations persist across all parameter scales (x-axis: Parameters in billions).
    \textbf{Left:} Average Model Performance on Numericals (y-axis: Exact Match score).
    \textbf{Center:} Average Model Performance on Numericals (y-axis: Tolerance-Based score).
    \textbf{Right:} Average Model Performance on MCQs (y-axis: Accuracy).
    }
    \label{fig:average_performance}
\end{figure*}

\subsection{Deduplication and Uniqueness Validation}
We conducted rigorous quality assurance: (1) Cross-benchmark deduplication with existing datasets using n-gram similarity analysis confirmed minimal overlap of only 6 questions (0.15\% of our dataset); (2) Advanced leakage detection using GPT-4o \cite{openai2024gpt4ocard} with four distinct methodologies: \textit{prefix completion testing} (systematic truncation at various points to detect completion patterns), \textit{paraphrasing detection} (generation of semantic equivalents), \textit{content modification analysis} (numerical and formula alterations), and \textit{reverse engineering} (concept-based generation); more details in supplementary. This comprehensive analysis identified a mere $\approx$ 8\% potential exposure; (3) $\mathcal{CP}_D$ questions from JEE (2020-2024) are inherently resistant to memorization due to multi-step reasoning requirements.

\subsection{Dataset Composition}

ChemPro contains 4,100 questions: $\mathcal{CP}_D$ (2,315, 56.4\%), $\mathcal{CP}_C$ (665, 16.2\%), $\mathcal{CP}_M$ (335, 8.2\%), and $\mathcal{CP}_E$ (795, 19.3\%). Each question underwent three-fold verification including source verification, expert review, and AI-assisted validation (detailed procedures in Figure \ref{fig:curation_process}).

\textbf{Difficulty distribution.} ChemPro intentionally contains more $\mathcal{CP}_D$ items to stress-test competitive, multi-step formulations; we therefore report results per-tier and per-subfield.

\textbf{Textual Adaptation.} Chemistry visuals were systematically converted to text using: standardized LaTeX equations, IUPAC nomenclature for structures, numbered sequences for reaction mechanisms, and structured descriptions for diagrams. These use representations commonly found in educational materials \citep{song2024resteemdatarepositoryeducational}, minimizing linguistic complexity while preserving visual provenance.

\subsection{Evaluation Framework and Experimental Setup}
 
For MCQs: $\text{Acc}_{\mathcal{M}} = \frac{1}{|\mathcal{M}|}\sum_{i=1}^{|\mathcal{M}|} \mathbb{I}[f(\mathcal{L}, q_i) = a_i]$\\

\noindent For numerical problems:
\begin{itemize}
\item \textbf{Exact Match}: $\text{Acc}_{exact} = \frac{1}{|\mathcal{N}|}\sum_{i=1}^{|\mathcal{N}|} \mathbb{I}[\hat{y}_i = y_i]$
\item \textbf{Tolerance}: $\text{Acc}_{tol} = \frac{1}{|\mathcal{N}|}\sum_{i=1}^{|\mathcal{N}|} \mathbb{I}[|\hat{y}_i - y_i| \leq \theta \cdot \frac{|y_i| + |\hat{y}_i|}{2}]$
\end{itemize}

where $\theta=0.1$ is the tolerance threshold. This approach distinguishes between conceptual understanding (tolerance-based) and computational precision (exact match).

\textbf{Answer Verification} Numericals are scored on the parsed numeric value extracted from \texttt{FINAL ANSWER} (not raw string equality). Questions specify integer / two-decimal rounding and units; if units are omitted, the expected SI-unit answer is in an appropriate range (details in supplementary).

\textbf{Model Selection.} We include 45+7 LLMs: (1) 40 open-source models from the OpenLLM Leaderboard representing top performers across five parameter scales (7B, 10B, 14, 32B, 70B) to capture scaling effects in general-purpose language models, 5 chemistry corpus pretrained and 2 latest geeral purpose releases; (2) State-of-the-art proprietary models: GPT-3.5-Turbo, GPT-4o, o1-mini, o3-mini, and o1 representing current commercial capabilities; (3) ChemCrow agentic framework for analysis vertical scaling with tool augmentation. Focus on general-purpose models addresses a critical gap: while specialized chemistry models proliferate for narrow research tasks, foundational chemistry remains underserved. Complete model list in Appendix.

\textbf{Evaluation Protocol.} All models evaluated pass@1. We probe COT, Self-Critique and ICL finalising an empirically robust system and wraparound prompts (separate for MCQs and Numericals) which are consistent for all models (system prompt is appended to user-input with wraparound prompt for reasoning models). ChemCrow is evaluated as-is.

\subsection*{Availability and Licensing}
The ChemPro benchmark is constructed from publicly available educational resources. NCERT textbooks are released under an open educational license by the Government of India, and JEE Mains examination papers are publicly distributed by the National Testing Agency. ChemPro will be released with full license compliance with all sources to ensure unrestricted access for research.
\section{Analysis and Discussion}

\begin{figure*}[t!]
    \includegraphics[width=0.33\linewidth]{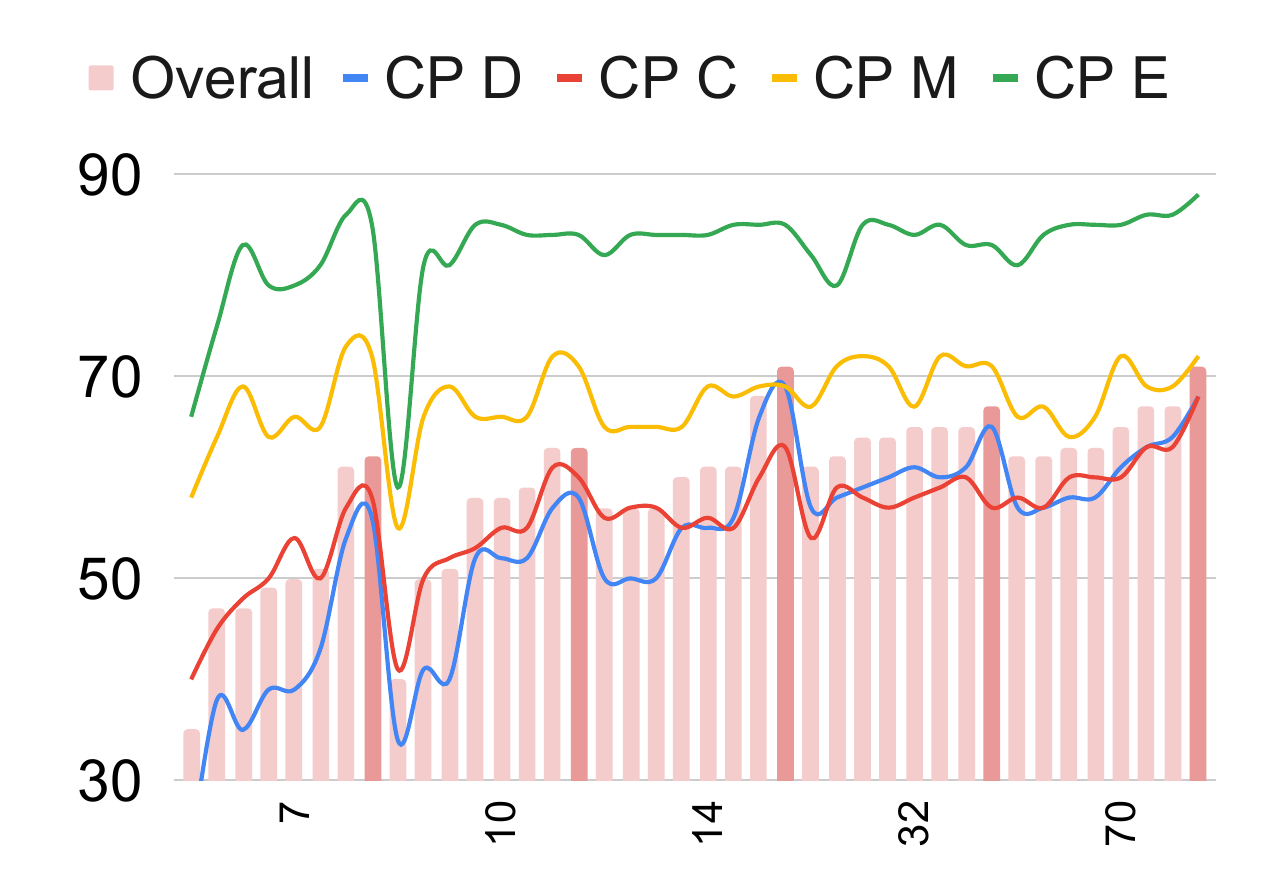}
    \includegraphics[width=0.33\linewidth]{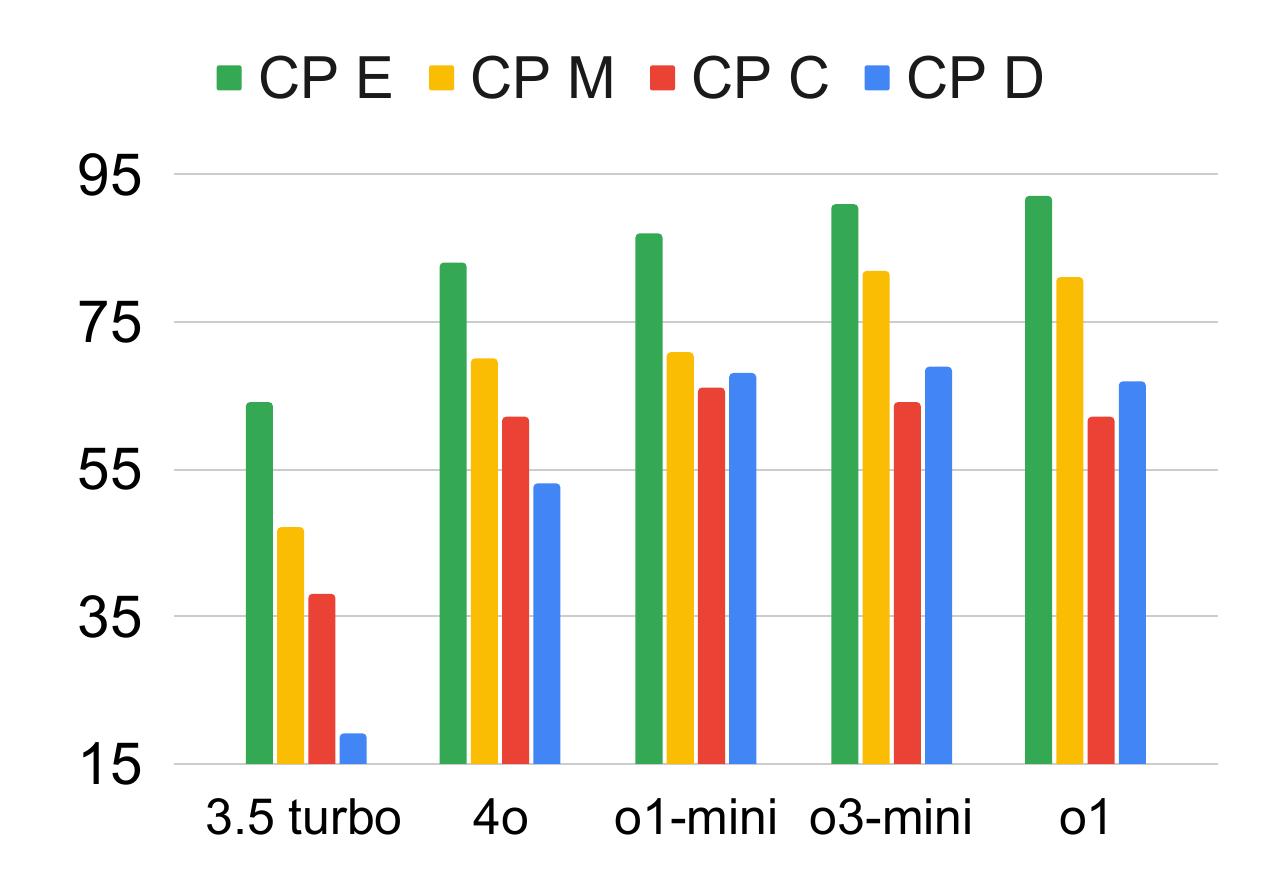}
    \includegraphics[width=0.33\linewidth]{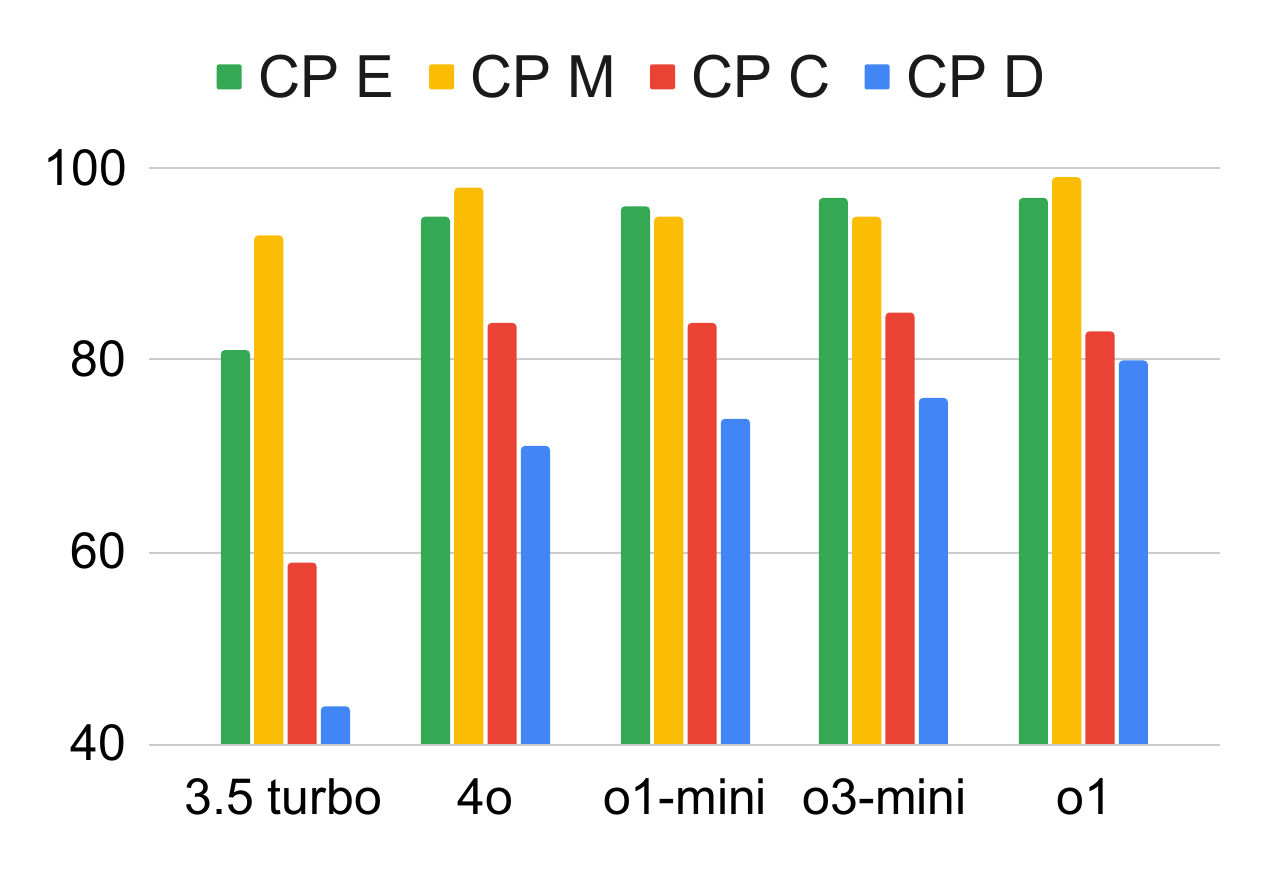}
    \caption{
    \textbf{\textit{Comprehensive Performance Analysis.}}
    \textbf{Left:} Overall tolerance-based performance across all model sizes showing scaling limitations.
    \textbf{Center:} OpenAI model performance on numerical problems across ChemPro sections (y-axis: Tolerance-Based score).
    \textbf{Right:} OpenAI model performance on multiple-choice questions across ChemPro sections (y-axis: Accuracy).
    }
    \label{fig:overall_performance}
\end{figure*}

\subsection{Human Performance Comparison and Evaluation}

Our human performance reference draws from established performance data on JEE Mains and NCERT board examinations from 2020--2024, the same period from which ChemPro questions are sourced. Historical data from these examinations shows that top 100 students typically score 97--100\% (refer Figure \ref{fig:human_comparison}) on chemistry sections. We emphasize that this is an \emph{indirect proxy}: the cited cohort did not take ChemPro end-to-end, and the exact ChemPro sampling may not match any single historical paper. We therefore use this reference primarily as an educationally grounded upper-bound context for expected mastery of the underlying curriculum, rather than as a controlled human-vs-model head-to-head evaluation. This comparison is analogous to AlphaGeometry's \citep{AlphaGeometryTrinh2024} silver medal performance evaluation on International Mathematics Olympiad.

\textbf{Significance}: ChemPro tests whether broadly-deployed LLMs can handle chemistry material students are expected to master. In addition to reporting accuracy, we use the progressive design to diagnose where robustness breaks as problems require longer chains of operations (e.g., conversions and cross-condition integration). Table \ref{tab:chemcrow_comparison} further suggests that tool-augmented agentic frameworks alone do not remove degradation at higher tiers.

\begin{table}[ht]
\centering
\small
\caption{\textbf{Performance Comparison} between Agentic Framework ChemCrow and GPT-4o on ChemPro MCQs}
\begin{tabular}{|l|cccc|}
\toprule
\textbf{System} & \textbf{$\mathcal{CP}_E$} & \textbf{$\mathcal{CP}_M$} & \textbf{$\mathcal{CP}_C$} & \textbf{$\mathcal{CP}_D$} \\
\midrule
ChemCrow & 97 & 98 & 84 & 68 \\
GPT-4o & 95 & 98 & 84 & 71 \\
\bottomrule
\end{tabular}
\label{tab:chemcrow_comparison}
\end{table}

\subsection{Empirical Evidence for Articulation Effects}

\textbf{Complex articulation}-defined by multi-step reasoning, unit conversions, and conceptual integration degrades performance across curriculum-aligned content. Importantly, we do not equate articulation complexity with surface linguistic length; rather, it reflects the structure of required operations (chained steps, conversions, and cross-condition integration).

ChemPro reveals this pattern (Figure \ref{fig:average_performance} \& Figure \ref{fig:overall_performance}) through performance measurement across difficulty sections. The progression from $\mathcal{CP}_E$ to $\mathcal{CP}_D$ represents increasing difficulty within the same curricular scope, as evidenced by: 1. \textbf{Consistent Performance Degradation}: All 45 evaluated models show declining accuracy as section difficulty increases, with an average $\approx$ 21-percentage-point drop from elementary to competitive formulations. 2. \textbf{Pattern Across Architectures}: This degradation pattern appears across diverse model architectures (gemma, Qwen, Falcon, PHI, etc.)\cite{gemmateam2024gemmaopenmodelsbased,qwen2025qwen25technicalreport, almazrouei2023falconseriesopenlanguage, phi4} and scales (7B to 70B+), indicating a systematic challenge rather than architecture-specific limitations. 3. \textbf{Preserved Relative Rankings}: While absolute performance varies, the relative difficulty ordering remains consistent across models, confirming that articulation complexity represents a measurable but under-attended dimension of challenge. This effect is pronounced for numericals, where unit handling and multi-step calculations are frequent, and errors can cascade.

\begin{figure}[ht!]
    \centering
    \includegraphics[width=0.98\linewidth]{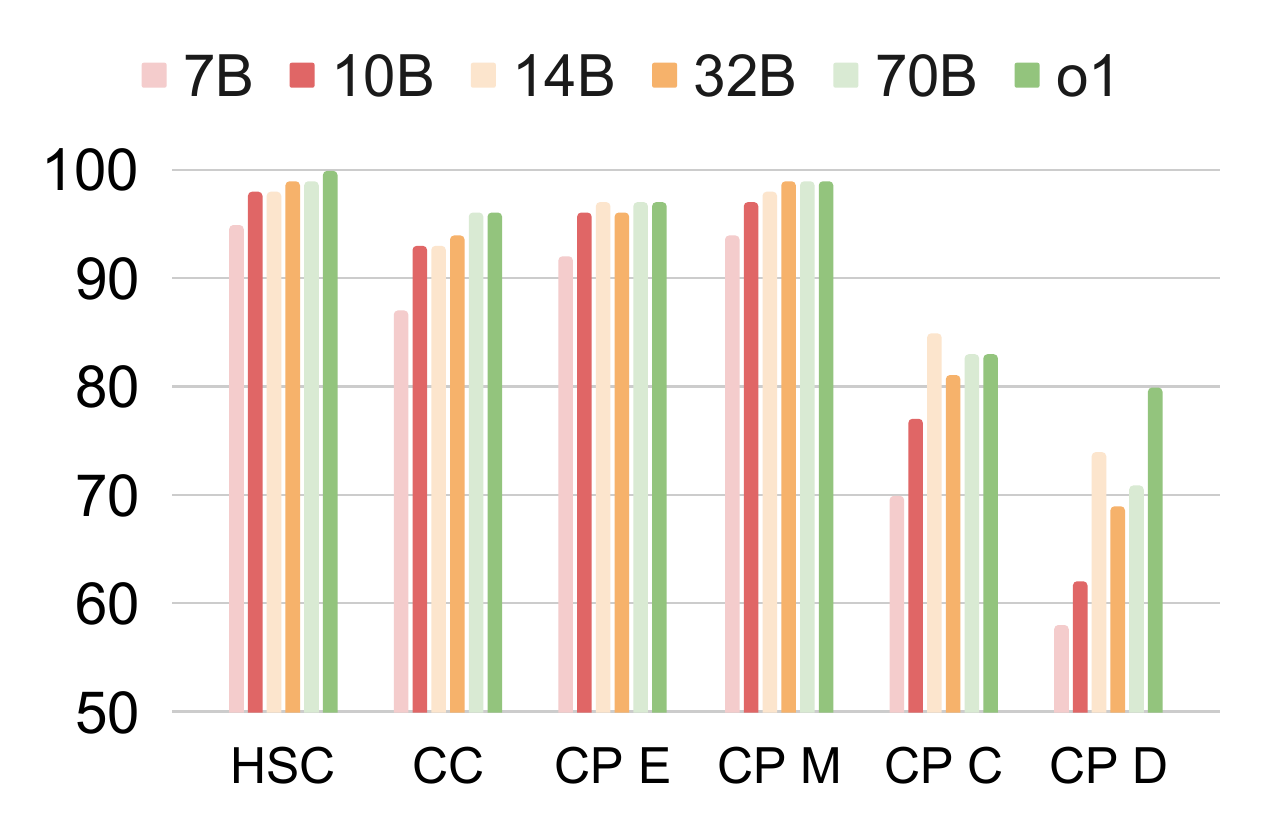}
    \caption{\textbf{\textit{Dataset Comparison:}} Performance of top models from each size category on ChemPro MCQs, College Chemistry (CC), and High School Chemistry (HSC). (x-axis: model sizes (7B to 70B \& Proprietary); y-axis: accuracy)}
    \label{fig:num_tolerance_perf}
\end{figure}

\begin{figure}[ht]
    \centering
    \includegraphics[width=0.95\linewidth]{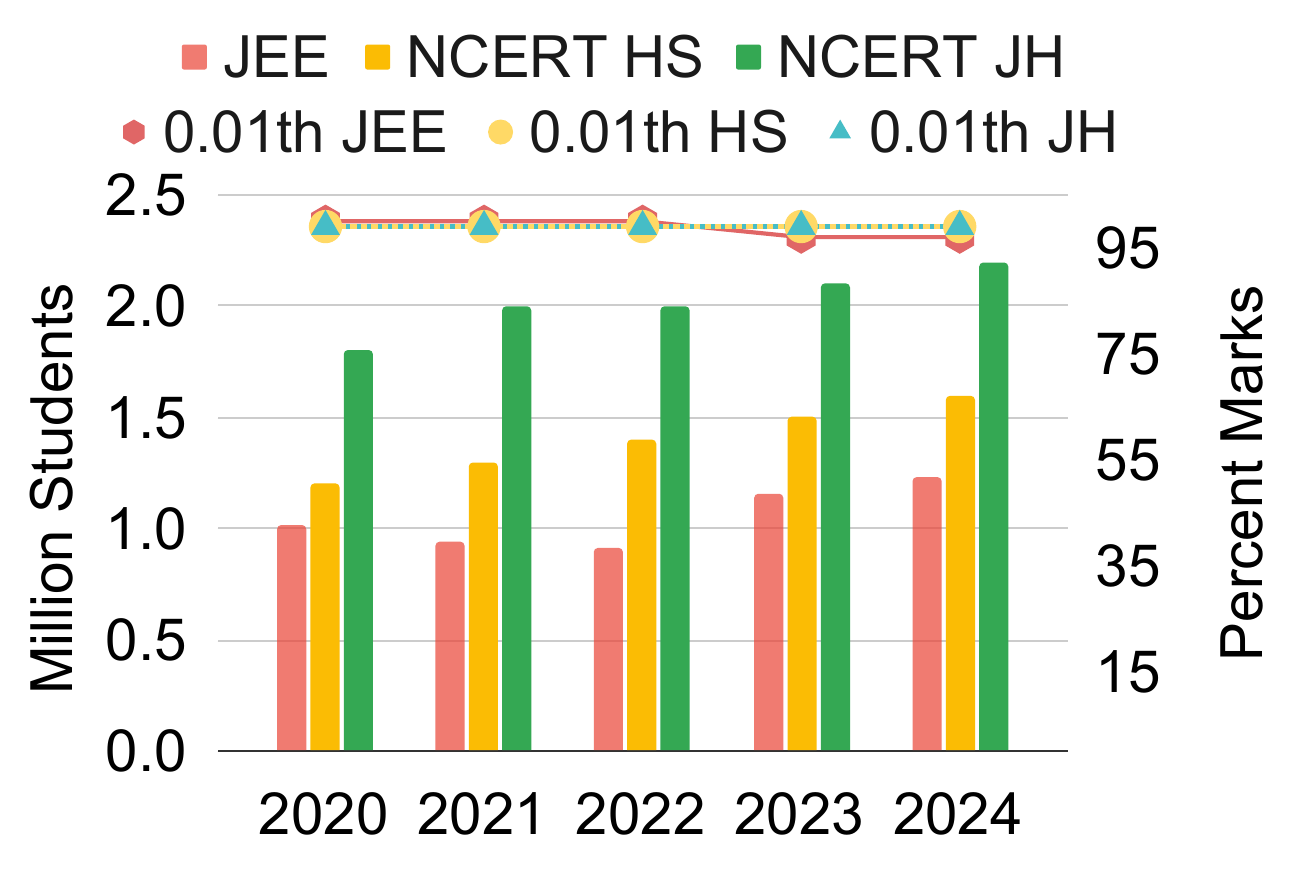}
    \caption{\textbf{\textit{Human vs. LLM Performance Comparison.}} Human performance on corresponding educational assessments significantly exceeds current LLM capabilities.}
    \label{fig:human_comparison}
\end{figure}
\vspace{-1em}

\textbf{Articulation and Conceptual Complexity}: Our curriculum-aligned design enables this distinction through educational provenance. The key insight is that $\mathcal{CP}_C$ and $\mathcal{CP}_D$ (JEE Mains) operate within identical curriculum boundaries, JEE Mains officially adheres to NCERT syllabus as defined by the National Testing Agency. Questions within each chemistry subdomain (biochemistry, organic, inorganic, physical chemistry) therefore assess the same conceptual scope while varying in formulation complexity. The systematic nature of performance degradation between these curriculum-equivalent sections (13-point average drop), consistent across models, supports the interpretation that articulation patterns contribute to difficulty due to composition rather than missing topic coverage.
\vspace{-0.5em}
\subsection{The Open-Source Capability Gap}
\textbf{Small Model $\approx$ 7B \& 10B}: Small models degrade sharply with competitive formulations, with numerical reasoning as a recurring failure mode. \textbf{Medium Model $\approx$ 14B \& 32B}: Medium models improve on easy tiers but remain unreliable as formulation complexity increases. \textbf{Large Models $\approx$ 70B}: Even the strongest open models remain meaningfully below educational reliability, especially on numericals.

\noindent Overall, current open models do not reliably master foundational chemistry under competitive articulation.

\begin{figure*}[t!]
    \includegraphics[width=0.33\linewidth]{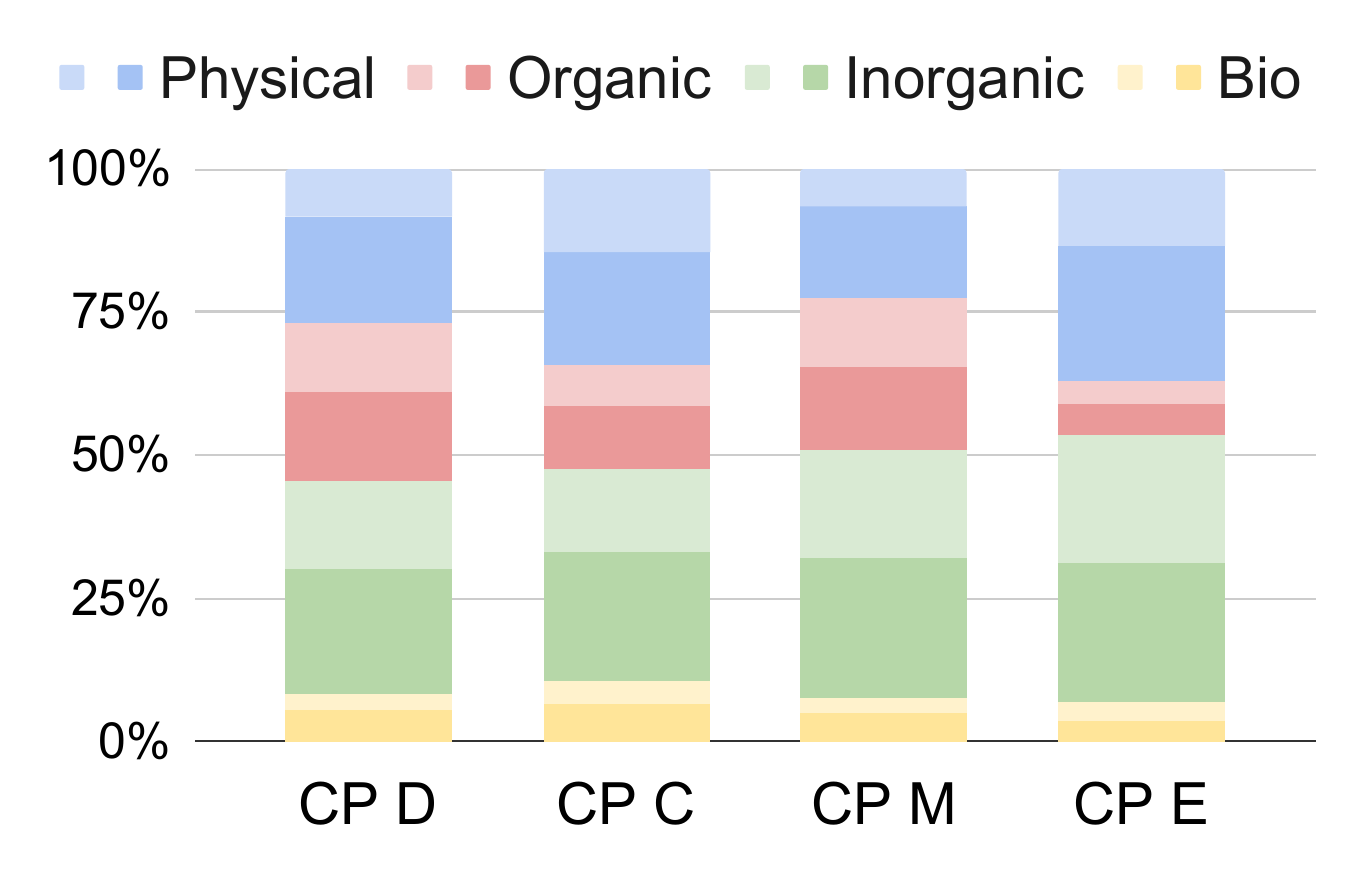}
    \includegraphics[width=0.33\linewidth]{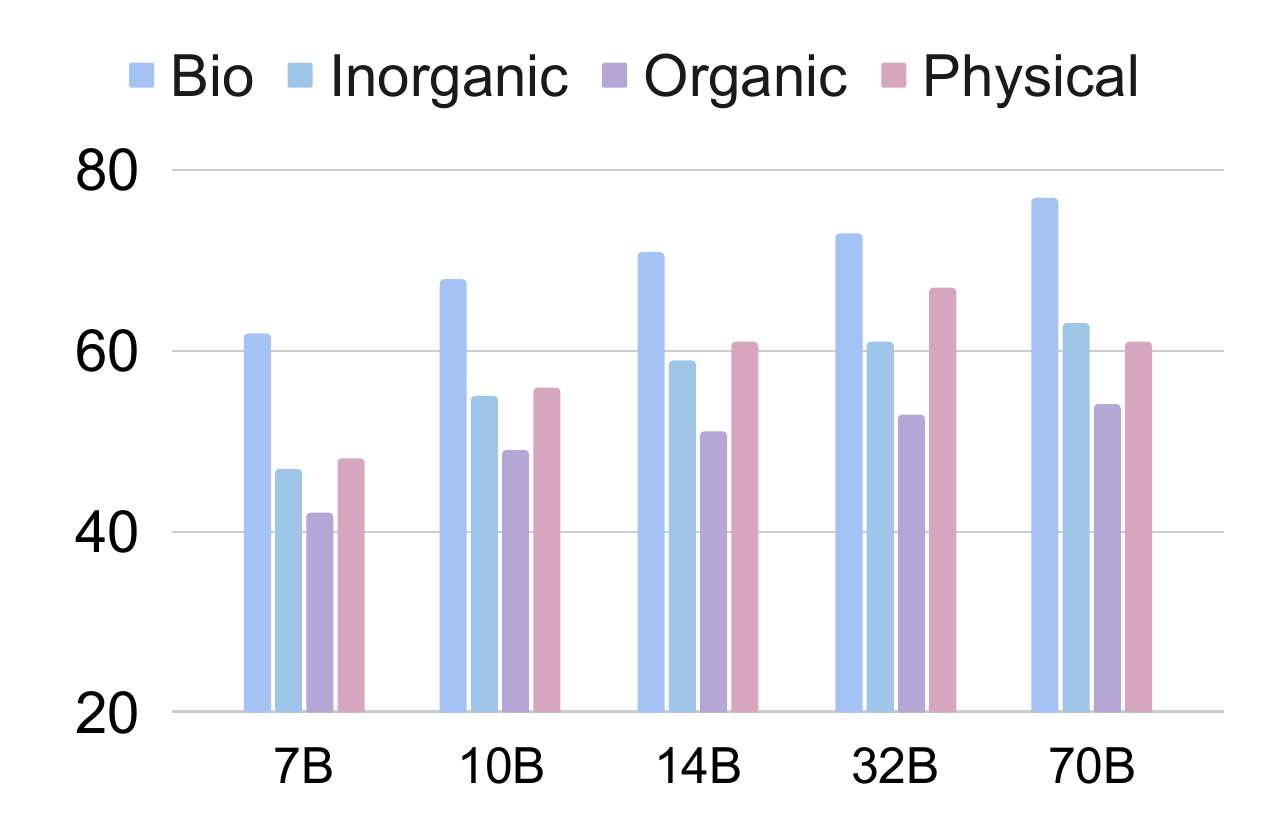}
    \includegraphics[width=0.33\linewidth]{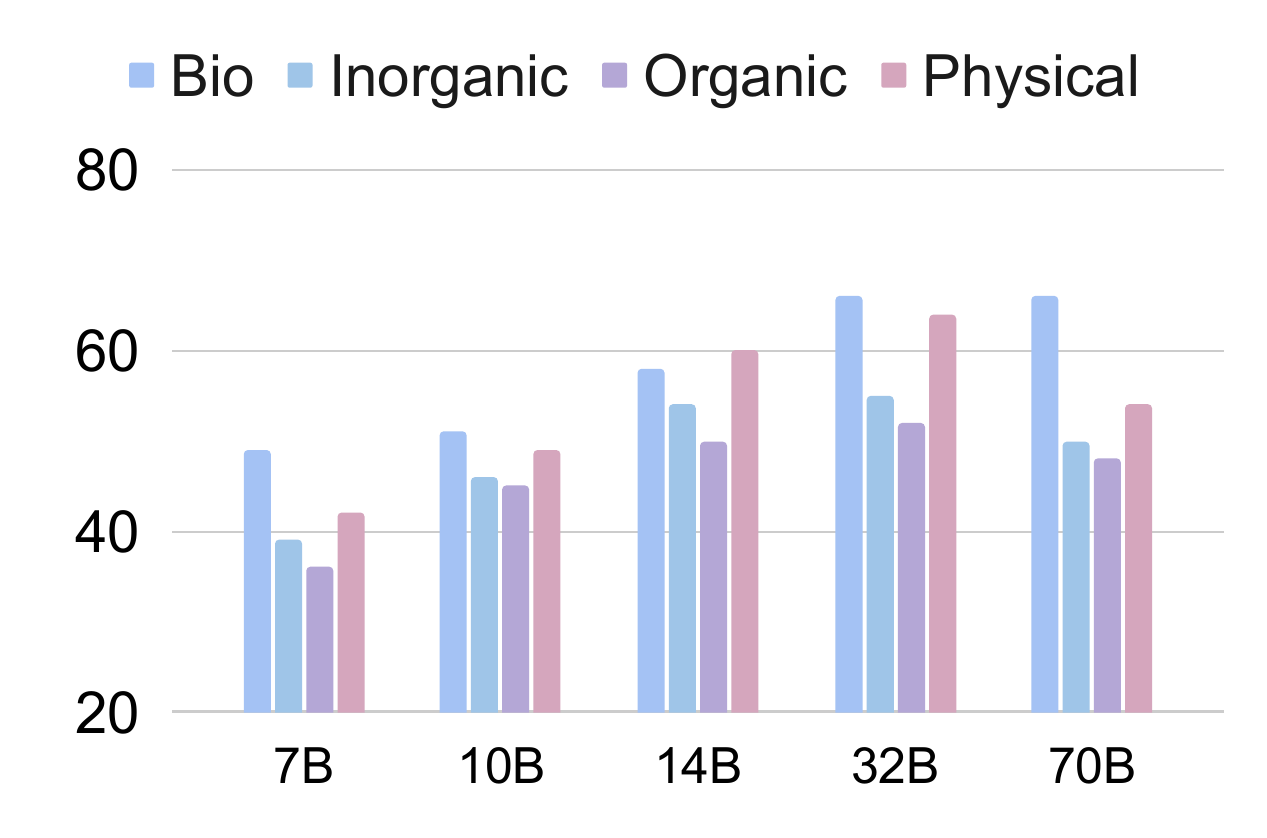}
    \caption{
    \textit{\textbf{Subfield Performance Analysis}}
    \textbf{Left:} Section-wise distribution of questions into subfields (y-axis: Dataset distribution; lighter shades are MCQs and darker shades are Numericals).
    \textbf{Center:} Model performance on multiple choice questions (y-axis: accuracy ) across subfields.
    \textbf{Right:} Model performance on numericals (y-axis tolerance-based score) across subfields.
    }
    \label{fig:subfield_analysis}
\end{figure*}

\subsection{Performance Scaling Analysis}

Analysis shows that parameter scaling is not robust to academic complexity: performance drops substantially with ChemPro tiers, and numerical limitations persist even for the largest models. Scaling improves performance on easier tiers but exhibits diminishing returns as questions demand longer procedural chains, indicating that robustness to articulation complexity remains a key bottleneck.

\section{Results}
We structure our findings around four key observations:

\noindent \textbf{Articulation as a Fundamental Challenge:}
Complex articulation represents a fundamental and systematic challenge for current LLMs, persisting across all tested model scales and architectures. Performance consistently degrades as questions progress from elementary formulations ($\mathcal{CP}_E$) to sophisticated multi-step reasoning requirements ($\mathcal{CP}_D$) within the same curricular scope. This pattern, marked by significant percentage-point performance drops, correlates with increased demands for multi-step reasoning, and conceptual integration, confirming that challenge lies in the reasoning process, with underlying scientific concepts.

\noindent \textbf{Diminishing Returns with Model Scaling:}
While larger models outperform smaller variants on foundational tasks-consistently achieving $\geq 90\%$ accuracy on $\mathcal{CP}_E$ MCQs-all models exhibit similar degradation patterns as formulation complexity increases (Figure \ref{fig:average_performance}). We observe performance convergence in the intermediate $\mathcal{CP}_C$ and $\mathcal{CP}_M$ sections. Even top-performing proprietary systems (o1, o3-mini), which achieve 74-76\% accuracy on competitive questions, follow the same fundamental degradation pattern. For comparison, 7-10B models achieve 53-67\% on these questions, with 70B+ variants reaching 68-71\%. These patterns strongly suggest that parameter scaling alone is insufficient to overcome multi-step scientific reasoning.

\noindent \textbf{Ineffectiveness of Current Agentic Frameworks:}
Our assessment of ChemCrow, reveals that current sophisticated reasoning frameworks do not overcome the identified limitations. ChemCrow provides only marginal improvements on elementary and intermediate tasks ($\mathcal{CP_E}$ and $\mathcal{CP_M}$) and fail to bridge the complexity gap for advanced problems ($\mathcal{CP_C}$ and $\mathcal{CP_D}$). For instance, it achieves performance comparable to its base GPT-4o model (Table \ref{tab:chemcrow_comparison}), indicating that access to tools and prompting strategies does not fundamentally resolve the underlying reasoning bottlenecks.

\noindent \textbf{Subfield-Specific Reasoning Bottlenecks:}
A detailed (Figure \ref{fig:subfield_analysis}) analysis reveals distinct and systematic bottlenecks across chemistry subfields. \textbf{Biochemistry} yields the highest accuracy but are often hindered by computational demands. \textbf{Organic Chemistry} shows severe performance drops on problems requiring advanced spatial reasoning and multi-step synthesis. In \textbf{Physical Chemistry}, models leverage mathematical formulations but frequently fail on precise numerical calculations and unit conversions. Finally, \textbf{Inorganic Chemistry} displays high variability, with unpredictable performance, indicating a fragile understanding of bonding, coordination chemistry and reactivity principles.

\noindent Our Evaluations on 7 additional (new) open-sourced models Table \ref{tab:model_names_additional} (3 General Purpose and 4 Chemistry Focused) has been provided in Table \ref{tab:new_models_performance}(Supplementary). The scores from these models re-verify our inferences and thereby establish a strong need for \textbf{ChemPro} benchmark and research oriented towards robustness against complex articulation. 
\section{Conclusions}

We introduce ChemPro, a novel curriculum-aligned progressive benchmark designed to rigorously evaluate and diagnose LLM capabilities in scientific reasoning. Our comprehensive evaluation of models unequivocally demonstrates a systematic pattern of performance degradation directly correlating with question articulation complexity. This consistent decline, observed across all models regardless of architecture or scale, exposes fundamental challenges in current language modeling approaches to multi-step scientific reasoning.

\noindent The findings from ChemPro are critical: Current architectural paradigms face inherent limitations in complex scientific reasoning that cannot be overcome solely through parameter scaling or even sophisticated agentic frameworks. ChemPro's curriculum-aligned progression validates that these observed failures occur on material students are expected to master, confirming genuine reasoning limitations rather than a lack of esoteric expert knowledge. By systematically profiling reasoning depth and identifying subfield-specific bottlenecks, ChemPro serves as a powerful diagnostic tool and provides a clear path forward for developing next-generation LLMs capable of robust and reliable scientific problem-solving, beyond superficial understanding to true conceptual mastery essential for educational and research applications.
{   \clearpage
    \bibliography{aaai2026}
}

\clearpage
\section*{Supplementary Material}
\label{supp}

\vspace{12pt}

\subsection*{Detailed Methodology}

\subsubsection*{Model Selection Criteria}

Model selection process involved: (1) Top 15 models on the OpenLLM Leaderboard based on aggregate performance across SOTA benchmarks (IFEval, BBH, GPQA, MMLU, MATH, MUSR), ensuring representation of the strongest available models within each parameter class; (2) Coverage of five parameter scales (7B, 10B, 14B, 32B, 70B+) to enable systematic scaling analysis; (3) Inclusion of diverse architectures (Llama, Qwen, Falcon, PHI) to assess architectural effects; (4) Integration of latest reasoning systems (o1, o1-mini, o3-mini) and agentic frameworks (ChemCrow) for comprehensive evaluation.

\subsubsection*{Evaluation Protocol Details}

All models were evaluated pass@1 with average of 5 runs; Token budget (LLM:8000; LRM:10000); Temperature(LLM:0.3; LRM:1); Top P(0.9). Numerical problems were scored using both exact match and tolerance-based metrics ($\theta = 0.1$) to differentiate between conceptual understanding and computational precision. Final answer extraction used a dynamic, regex-based parsing of the \texttt{FINAL ANSWER} field: for numericals we extract the numeric value (handling scientific notation and common formatting), and scoring is performed on the resulting numeric value rather than raw string equality. To reduce ambiguity, numerical questions specify that the final answer should be an integer or a numerical value rounded to two decimal places, with the relevant units specified; when units are not explicitly requested, questions are formulated such that the expected SI-unit answer is in a clear, appropriate range. \{Machine: Nvidia A100 80GB x 2\}

\subsubsection*{Advanced Leakage Detection Methodology}

To robustly assess potential model leakage from training data, we employed a four-pronged analytical framework using GPT-4o. Each method targets a different dimension of memorization detection, enabling a comprehensive evaluation across question types and sources.

\begin{enumerate}
    \item \textbf{Prefix Completion Testing}\\
    This method involves systematic truncation of input questions to various prefix lengths, followed by model probing to observe completion consistency. Let $Q = \{w_1, w_2, \ldots, w_n\}$ be a tokenized question.\\
    We define a set of truncated prefixes:
    \[
        Q_k = \{w_1, w_2, \ldots, w_k\}, \quad \text{for } k = 1, 2, \ldots, n-1
    \]
    We evaluate whether the model completes $Q_k$ to approximate the original suffix $\{w_{k+1}, \ldots, w_n\}$. A high cosine similarity $\cos(\theta)$ between embeddings of the generated continuation and the ground-truth suffix indicates potential memorization.

    \item \textbf{Semantic Paraphrasing Detection}\\
    We generate semantic equivalents $\hat{Q}$ of the original question $Q$ using LLM-based paraphrasers and back-translation. The model is then prompted with $\hat{Q}$:
    \[
        \text{sim}(Q, \hat{Q}) > \tau \Rightarrow \text{valid paraphrase}
    \]
    The response is compared against the original solution. Near-identical solutions across paraphrases indicate potential exposure through indirect memorization.

    \item \textbf{Content Modification Analysis}\\
    This method probes whether the model understands generalized principles or merely memorized specific data. Numerical constants, units, or formulae in $Q$ are systematically perturbed to yield $Q'$, e.g.:
    \[
        \text{Original: } F = ma; \quad \text{Modified: } F = 2ma
    \]
    Let $S_Q$ and $S_{Q'}$ be the model's solutions to $Q$ and $Q'$, respectively. We compute semantic and numerical deltas:
    
    $\Delta = \lVert S_Q - S_{Q'} \rVert$,
    A low $\Delta$ despite semantic divergence suggests general understanding; a high $\Delta$ with minor content changes suggests rote memorization.

    \item \textbf{Reverse Engineering via Conceptual Abstraction}\\
    In this method, we abstract core concepts from target questions (e.g., conservation laws, reaction kinetics) and regenerate novel questions $Q^*$ not present in known datasets. These are used to query the model:
    \[
        Q^* \sim \text{conceptual basis of } Q, \quad Q^* \notin \mathcal{D}_{\text{train}}
    \]
    High similarity in responses across $Q$ and $Q^*$, or spontaneous recognition of question structure in $Q^*$, may signal latent memorization from concept-rich samples.

\end{enumerate}

Across all methodologies, we identified approximately \textbf{8\% potential exposure} suggestive of memorization artifacts.

\subsection*{Figure 1 Axis Derivation Details}
The bubble chart in Figure 1 (Left) visualizes benchmarking attributes across three dimensions:
\noindent \textbf{Y-axis (LLM Difficulty):} Operationalized as the mean multiple-choice question (MCQ) accuracy across all evaluated LLMs on each respective benchmark. Accuracy values are inverted (i.e., lower accuracy corresponds to a higher position on the y-axis) to reflect greater empirical difficulty for language models.

\noindent \textbf{X-axis (Academic Succession):} Represents the ordinal progression of educational curricula (Elementary, Middle School, High School, Undergraduate, Graduate/Expert). The exact x-coordinates were determined through expert review combined with source provenance. Although academic stages are nominally discrete, the axis is plotted as a continuum because real-world curricula exhibit continuous overlap. For instance, elementary chemistry concepts ($\mathcal{CP}_E$) preceed and overlap with junior high curricula (6th to 8th grade) where foundational material often repeats, which in turn expands into high school (9th to 12th grade). The continuous x-coordinates for ChemPro's tiers and external benchmarks reflect this graduated, overlapping nature of educational progression.

\noindent \textbf{Bubble Size:} Proportional to the total number of questions.

\subsection*{Detailed Experimental Results (Tables)}

\begin{table}[ht]
\centering
\small
\caption{\textbf{Performance by Subfield - MCQ Accuracy (\%)}}
\begin{tabular}{|l|cccc|}
\toprule
\textbf{section} & \textbf{Biochem.} & \textbf{Inorg.} & \textbf{Org.} & \textbf{Phys.} \\
\midrule
$\mathcal{CP}_E$ & 98.7 $\pm$ 1.2 & 90.5 $\pm$ 4.3 & 88.2 $\pm$ 3.8 & 85.5 $\pm$ 4.1 \\
$\mathcal{CP}_M$ & 96.8 $\pm$ 2.1 & 84.4 $\pm$ 6.7 & 86.5 $\pm$ 5.2 & 73.5 $\pm$ 3.9 \\
$\mathcal{CP}_C$ & 92.7 $\pm$ 3.8 & 63.3 $\pm$ 7.8 & 58.5 $\pm$ 6.1 & 72.4 $\pm$ 5.3 \\
$\mathcal{CP}_D$ & 71.2 $\pm$ 8.1 & 59.1 $\pm$ 9.2 & 50.1 $\pm$ 7.4 & 56.7 $\pm$ 8.8 \\
\bottomrule
\end{tabular}
\label{tab:subfield_mcq_detailed}
\end{table}

\begin{table}[ht]
\centering
\small
\caption{\textbf{Performance by Subfield - Tolerance Match (\%)}}
\begin{tabular}{|c|llll|}
\toprule
\textbf{section} & \textbf{Bio} & \textbf{Inorganic} & \textbf{Organic} & \textbf{Physical} \\
\midrule
$\mathcal{CP}_E$ & 100.0 $\pm$ 0 & 94.1 $\pm$ 2.1 & 87.0 $\pm$ 4.6 & 79.3 $\pm$ 3.8 \\
$\mathcal{CP}_M$ & 98.4 $\pm$ 1.8 & 67.3 $\pm$ 8.9 & 75.1 $\pm$ 6.2 & 62.6 $\pm$ 4.7 \\
$\mathcal{CP}_C$ & 90.6 $\pm$ 5.2 & 50.4 $\pm$ 9.8 & 50.1 $\pm$ 7.9 & 65.8 $\pm$ 6.1 \\
$\mathcal{CP}_D$ & 60.0 $\pm$ 12.3 & 47.8 $\pm$ 11.7 & 45.5 $\pm$ 9.8 & 52.3 $\pm$ 10.2 \\
\bottomrule
\end{tabular}
\label{tab:subfield_num_detailed}
\end{table}

\noindent \textbf{Systematic Failure Patterns}:
\begin{enumerate}
\item \textbf{Multi-step Reasoning Breakdown}: Models frequently fail on problems requiring >3 sequential logical steps, even when individual steps are within their capabilities.
\item \textbf{Numerical Calculation Errors}: Persistent arithmetic mistakes in stoichiometry and equilibrium calculations, despite correct conceptual setup.
\item \textbf{Context Integration Failures}: Inability to synthesize information across problem statements, particularly in organic reaction mechanisms.
\item \textbf{Unit Conversion Errors}: Systematic mistakes in dimensional analysis and unit consistency checks.
\end{enumerate}

\begin{figure*}[t]
\centering
\begin{minipage}{0.9\textwidth}
\begin{center}
\subsection*{System/Instruction Prompt:}

\textbf{For MCQs:}\\
Given the multiple choice question, solve and return a crisp, concise, and concrete solution under the heading \texttt{SOLUTION:}. Then ONLY return the letter ( A/B/C/D ) of the correct choice under the heading \texttt{FINAL ANSWER:}.

\vspace{0.5em}

\textbf{For Numericals:}\\
Given the numerical question, solve and return a crisp, concise, and concrete solution under the heading \texttt{SOLUTION:}. Then ONLY return the correct answer (\$ final numerical value \$) under the heading \texttt{FINAL ANSWER:}.

\vspace{1em}

\textbf{Prompt Template:}
\begin{verbatim}
INSTRUCTION:
< System Prompt >

MCQ/NUMERICAL:
< Question Content >

OPTIONS: (In case of MCQs)
< Ordered Options >
\end{verbatim}
\end{center}
\end{minipage}
\end{figure*}

\begin{figure}[ht!]
    \centering
    \includegraphics[width=0.98\linewidth]{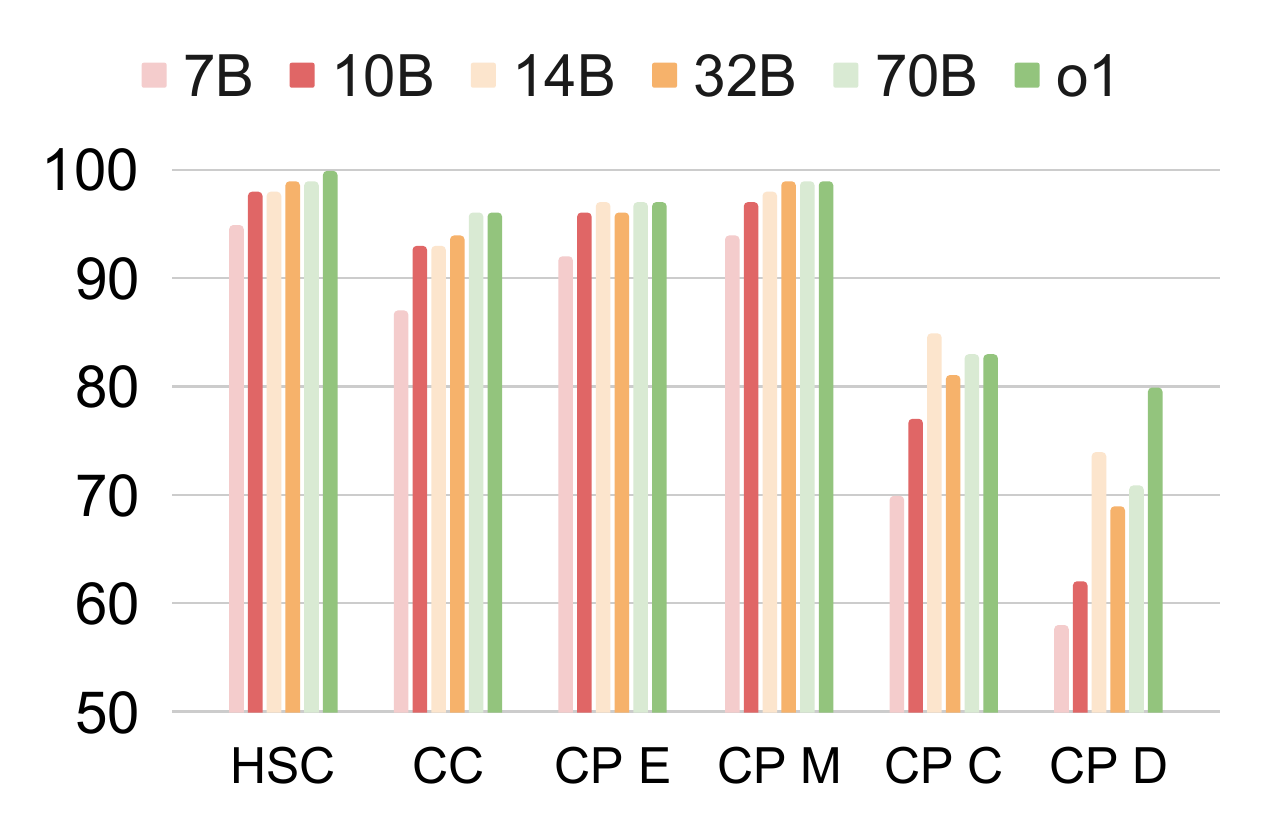}
    \caption{\textbf{\textit{Dataset Comparison:}} Performance of top models from each size category on ChemPro MCQs, College Chemistry (CC), and High School Chemistry (HSC). (x-axis: model sizes (7B to 70B \& Proprietary); y-axis: accuracy)}
\end{figure}

\begin{figure}[ht!]
    \centering
    \includegraphics[width=0.98\linewidth]{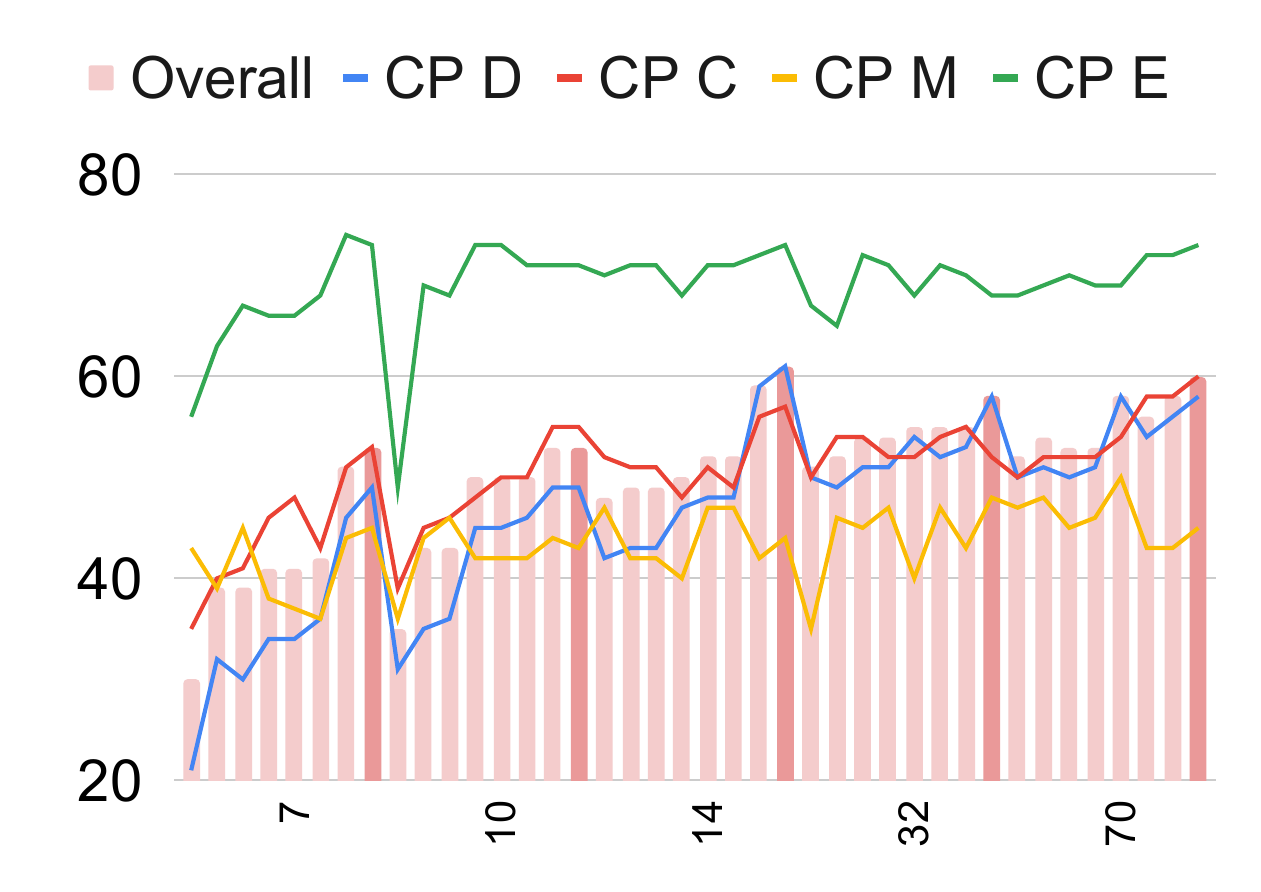}

    \caption{\textbf{\textit{All Performance on ChemPro Numericals:}} Performance of models across ChemPro Numericals by difficulty level ($\mathcal{CP}_D$: Difficult, $\mathcal{CP}_C$: Challenging, $\mathcal{CP}_M$: Medium, $\mathcal{CP}_E$: $\mathcal{CP}_E$) and overall accuracy. (x-axis: model sizes (7B to 70B); y-axis: exact match scores)}
    \label{fig:num_exact_perf}
\end{figure}

\begin{figure}[ht!]
    \centering
    \includegraphics[width=0.98\linewidth]{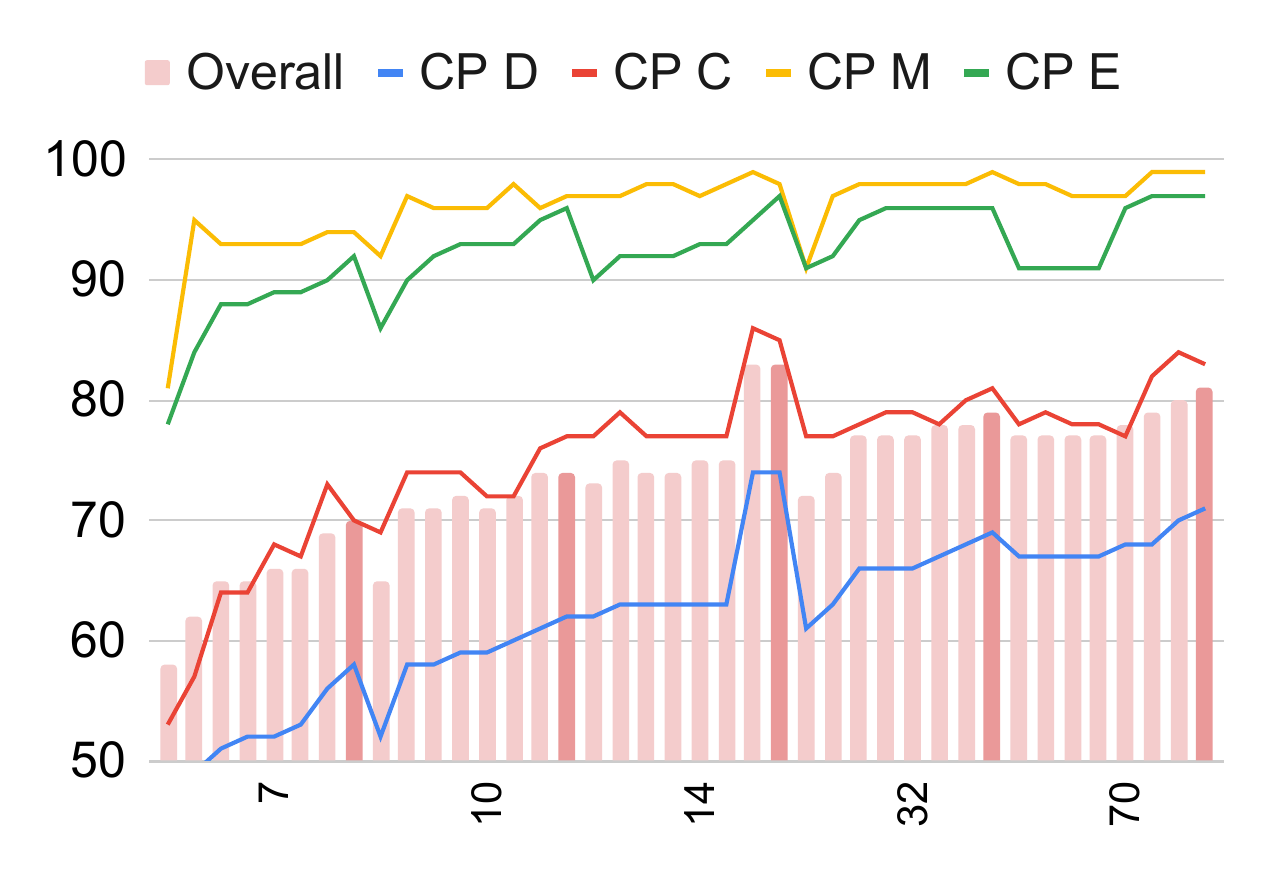}
    \caption{\textbf{\textit{Best Performance on MCQs:}} Best accuracy achieved by models of varying sizes (7B, 10B, 14B, 32B, and 70B) across ChemPro MCQ difficulty levels: $\mathcal{CP}_D$ (Difficult), $\mathcal{CP}_C$ (challenging), $\mathcal{CP}_E$ ($\mathcal{CP}_E$), and $\mathcal{CP}_M$ (Medium). Larger models consistently perform better, with accuracy increasing from $\mathcal{CP}_D$ to $\mathcal{CP}_E$ stages, highlighting the impact of model size on performance.}
    \label{fig:mcq_perf}
\end{figure}

\begin{figure}[ht!]
    \centering
    \includegraphics[width=0.98\linewidth]{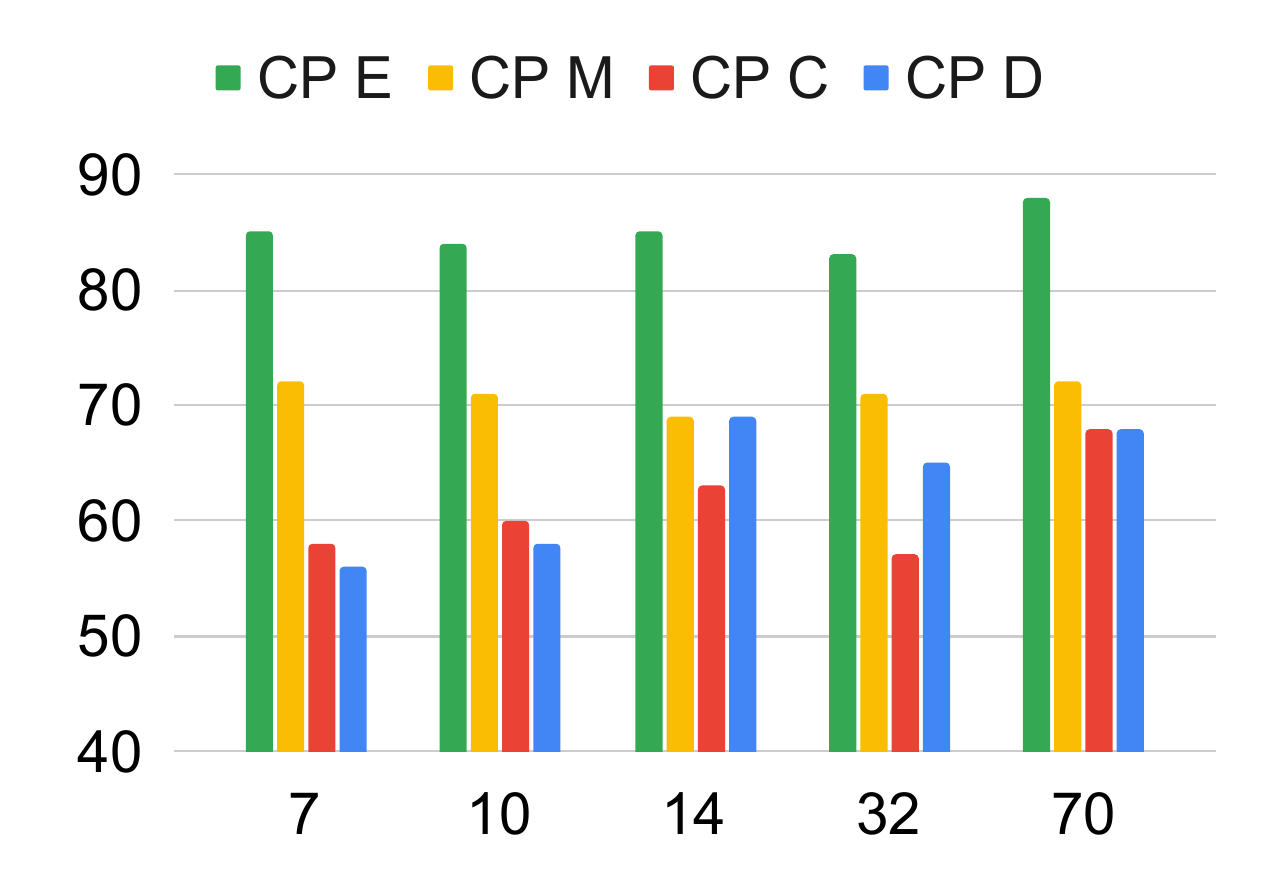}
    \caption{\textbf{\textit{Best Performance on Numericals:}} Best accuracy with tolerance achieved by models of varying sizes (7B, 10B, 14B, 32B, and 70B) across ChemPro Numericals difficulty levels: $\mathcal{CP}_D$ (Difficult), $\mathcal{CP}_C$ (challenging), $\mathcal{CP}_E$ ($\mathcal{CP}_E$), and $\mathcal{CP}_M$ (Medium). Larger models consistently perform better, with accuracy increasing from $\mathcal{CP}_D$ to $\mathcal{CP}_E$ stages, highlighting the impact of model size on performance.}
    \label{fig:best_tolerance_acc}
\end{figure}

\begin{table*}[ht]
\small
\caption{\textbf{\textit{LLM variants} (original)} evaluated on ChemPro benchmark.
}
    \label{tab:model_names_original}
    \begin{center}
    \begin{tabularx}{\linewidth}{|Y|Y|Y|}
        \toprule
        \rowcolor{shadecolor}\textbf{7B} & \textbf{10B} & \textbf{14B}\\
        \midrule
        Mawared-T1 & falcon3-10b-tensopolis-v1 & Phi-4-Model-Stock-v4\\
        Qwen2.5-7B-HomerCreative-Mix & virtuoso-lite-tensopolis-v1 & Luminis-PHI-4\\
        SJT-7B-V1.1 & virtuoso-lite-tensopolis-v2 & Lamarck-14B-v0.7\\
        KytheraMix-7B-V0.2 & MT3-Gen4-gemma-2-9B & Saka-14B\\
        Falcon3-Jessi-v0.4-7B-Slerp & MT-Merge4-gemma-2-9B & Lamarckvergence-14B\\
        HomerCreativeAnvita-Mix-Qw7B & Virtuoso-Lite & virtuoso-small-v2-tensopolis-v1\\
        Falcon3-7B-Instruct & Falcon3-10B-Instruct & Virtuoso-Small-v2\\
        QandoraExp-7B & falconthink3-10b-it & Awqward2.5-32B-Instruct\\
        \midrule
        \rowcolor{shadecolor}\textbf{32B} & \textbf{70+B} & \textbf{Proprietary}\\
        \midrule
        oxyge1-33B & ultima-72B & \\
        ultima-32B & CalmeRys-78B-Orpo-v0.1 & OpenAI GPT-3.5-Turbo\\
        PathFinderAi3.0 & Rombos-LLM-V2.5-Qwen-72b & OpenAI GPT-4o\\
       , Qwentile2.5-32B-Instruct & shuttle-3 & OpenAI GPT-4o (with ChemCrow)\\
        Rombos-LLM-V2.5-Qwen-32b & calme-2.4-rys-78b & OpenAI o1-mini\\
        Qwen2.5-32B-Instruct-abliterated-v2 & calme-3.1-instruct-78b & OpenAI o3-mini\\
        openbuddy-qwq-32b-v24.1-200k & calme-3.2-instruct-78b & OpenAI o1\\
        Awqward2.5-32B-Instruct & Homer-v1.0-Qwen2.5-72B & \\
        \bottomrule
    \end{tabularx}
    \end{center}
\end{table*}

\begin{table*}[ht]           
  \small            
  \caption{\textbf{\textit{LLM variants} (additional)} evaluated on ChemPro benchmark.  
  }      
      \label{tab:model_names_additional}  
      \begin{center}           
      \begin{tabularx}{\linewidth}{|Y|Y|Y|}          
          \toprule  
          \rowcolor{shadecolor}\textbf{7--8B} & \textbf{14B} & \textbf{20--32B}\\     
          \midrule  
          ChemDFM-v1.5-8B \citep{DFM} & ChemDFM-v2.0-14B \citep{DFM} & ChemLLM-20B-Chat-DPO \citep{zhang2024chemllmchemicallargelanguage}\\      
          ChemLLM-7B-Chat-1\_5-DPO \citep{zhang2024chemllmchemicallargelanguage} & Qwen3-14B  \citep{yang2025qwen3technicalreport} & Qwen3-32B \citep{yang2025qwen3technicalreport}\\    
          Qwen3-8B \citep{yang2025qwen3technicalreport} & & \\      
          \bottomrule          
      \end{tabularx}           
      \end{center}  
  \end{table*}

\begin{figure*}[ht]
    \centering

    \includegraphics[width=\linewidth]{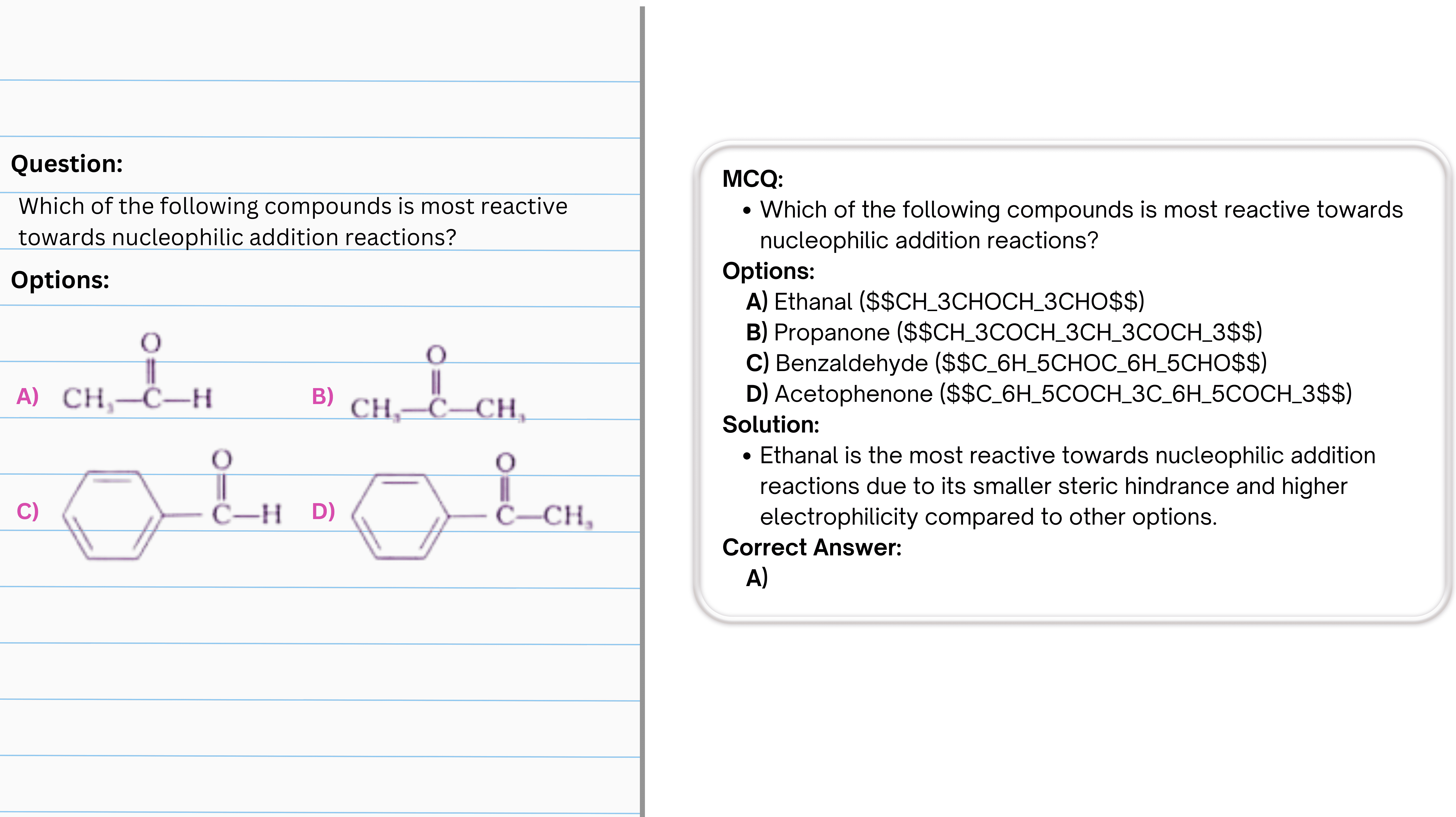}
    \caption{\textbf{\textit{Textual Adaptation Example 1:}} Adaptation in MCQs.}
    \label{fig:mcqcovert}
\end{figure*}
\begin{figure*}
    \centering

    \includegraphics[width=\linewidth]{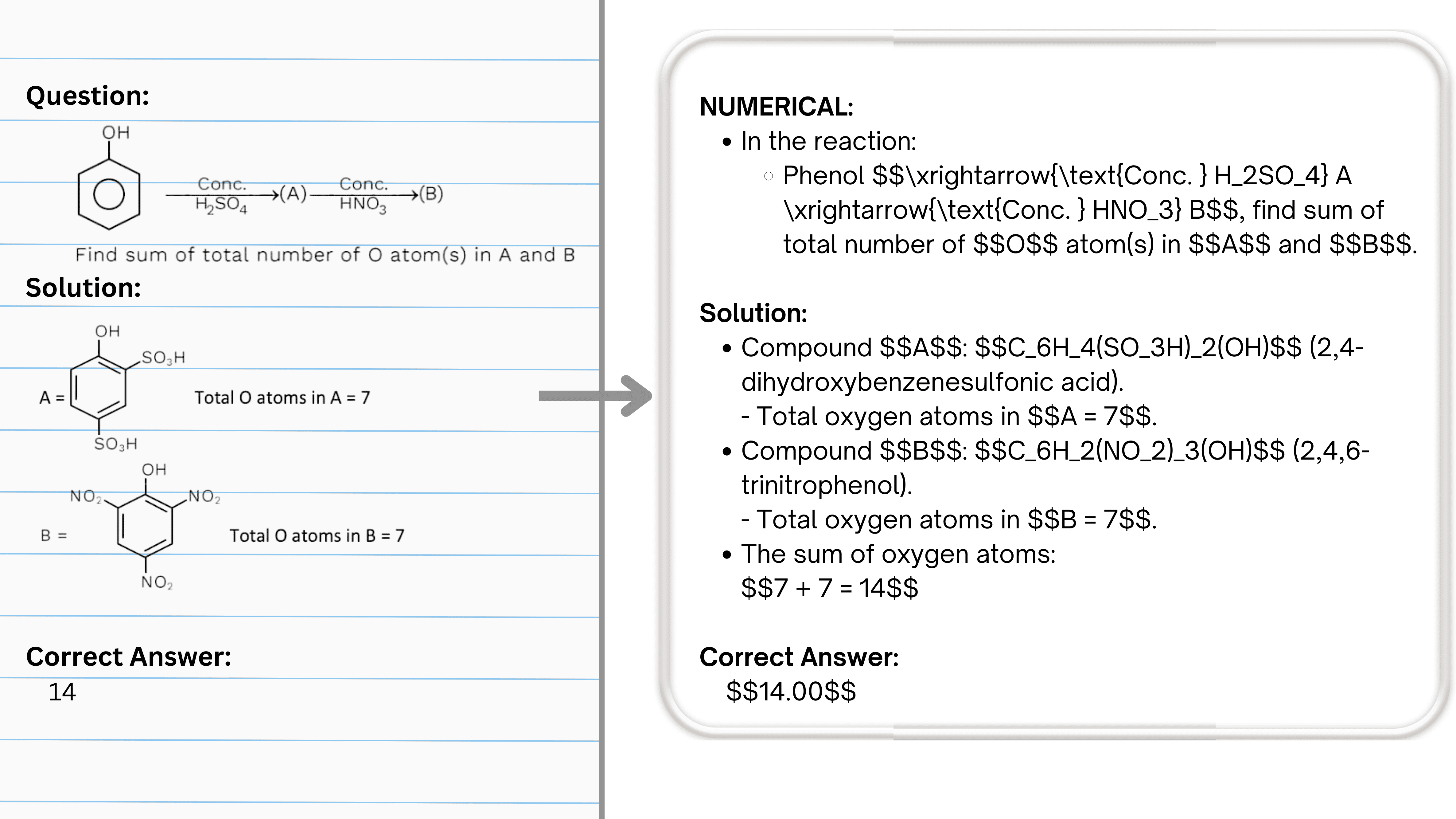}
    \caption{\textbf{\textit{Textual Adaptation Example 2:}} Adaptation in Numericals.}
    \label{fig:num_convert}

\end{figure*}

\begin{figure*}[ht]
    \centering
    \includegraphics[width=\linewidth]{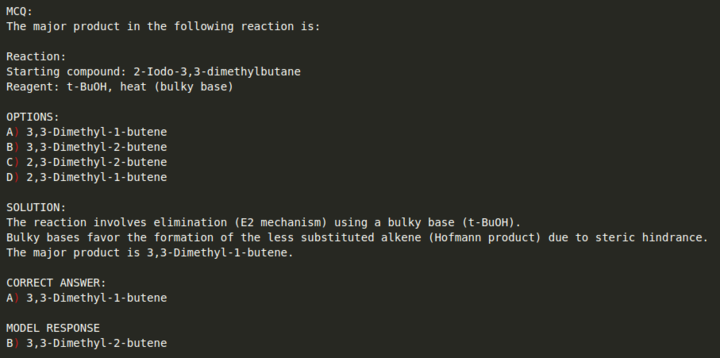}
    \caption{\textbf{\textit{Model Failure Example 1:}} This question requires multiple steps, complex interpretations, and self-memorization, which exceed the current capabilities of the LLM.}
    \label{fig:mcq_wrong_diff}
\end{figure*}

\begin{figure*}[ht]
    \centering
    \includegraphics[width=\linewidth]{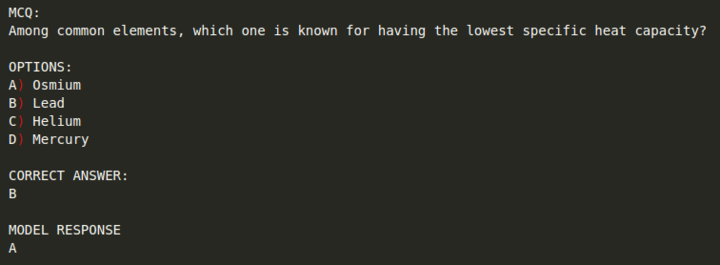}
    \caption{\textbf{\textit{Model Failure Example 2:}} The specific heat values for both osmium and lead are often reported as 0.13 J/g·K on the internet. However, Lead's actual specific heat value is 0.128 J/g·K, which highlights the LLM's incomplete knowledge leading to an incorrect answer in this case.}
    \label{fig:mcq_easy_wrong}
\end{figure*}

\begin{figure*}[ht]
    \centering
    \includegraphics[width=\linewidth]{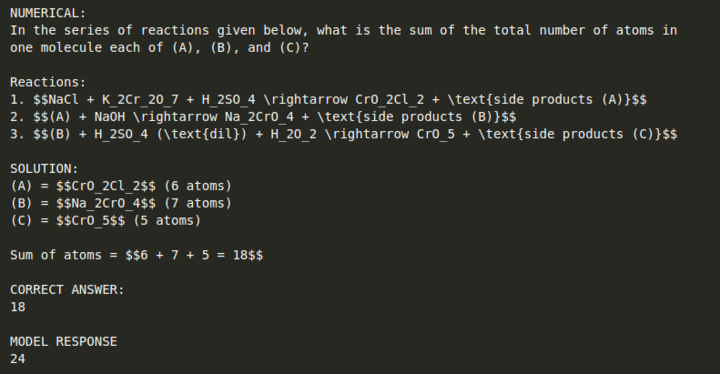}
    \caption{\textbf{\textit{Model Failure Example 3:}} This question requires multi-step reasoning, counting abilities, and domain-specific knowledge, which the LLMs lack.}
    \label{fig:num_wrong_diff}
\end{figure*}

\begin{figure*}[ht]
    \centering
    \includegraphics[width=\linewidth]{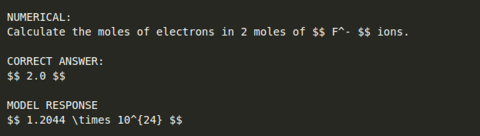}
    \caption{\textbf{\textit{Model Failure Example 4:}} The LLM appears to have altered the essence of a trick question and fallen into the intended trap.}
    \label{fig:num_wrong_easy}
\end{figure*}

\clearpage
\begin{figure*}[ht]
    \includegraphics[width=0.5\linewidth]{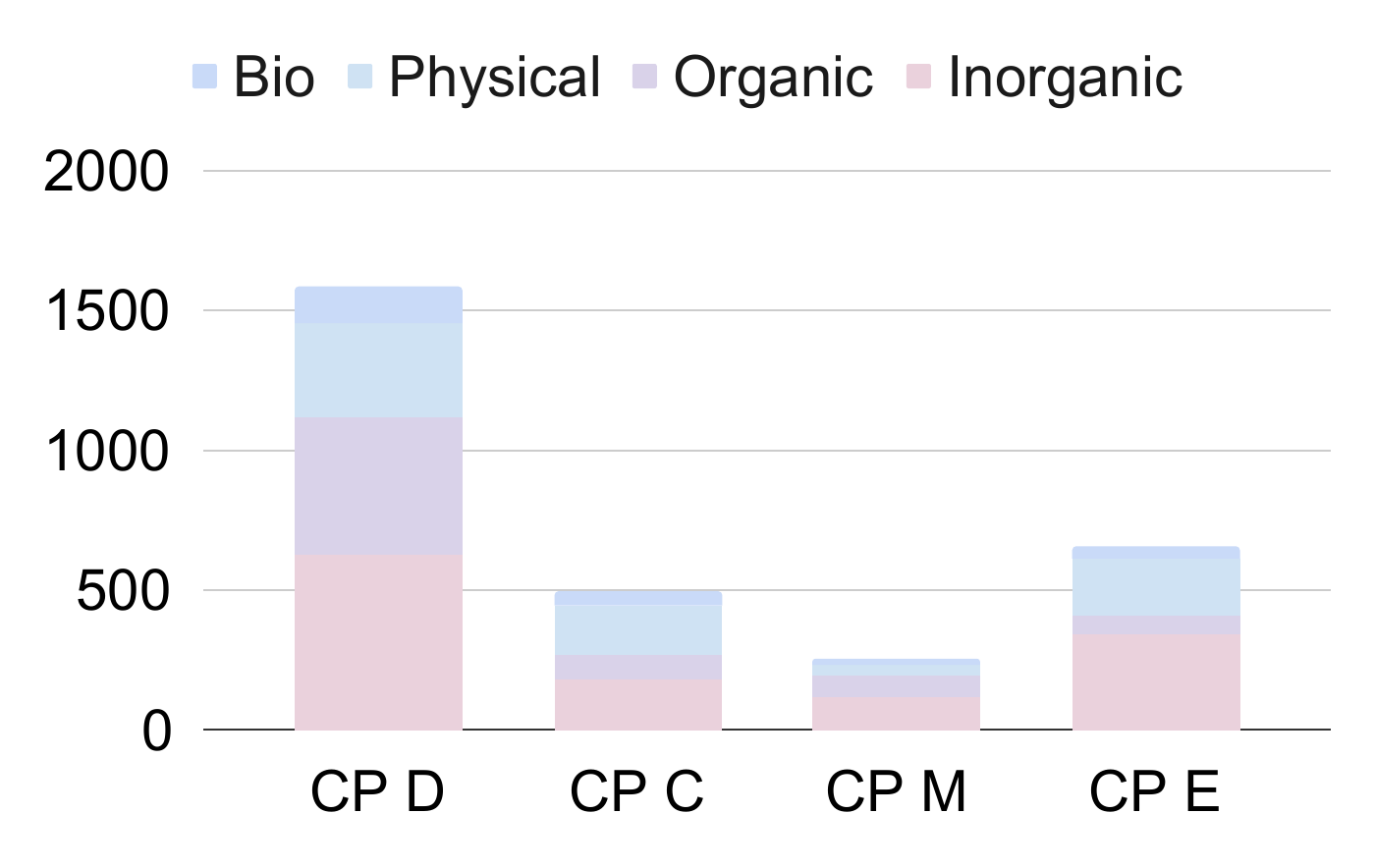} \quad
    \includegraphics[width=0.5\linewidth]{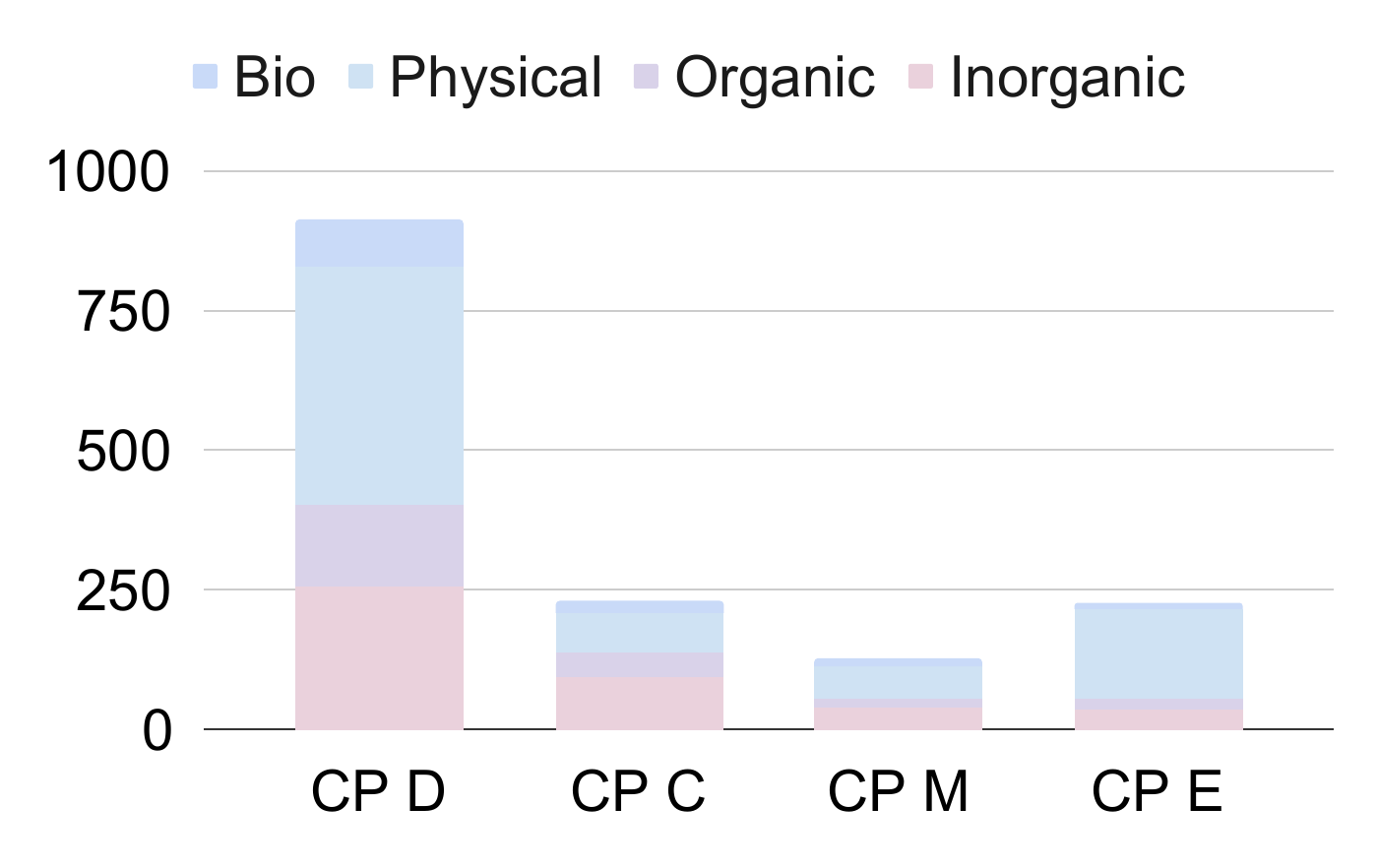}
    \caption{\textbf{\textit{Subfield Attribution :}} Subfield-wise distribution of ChemPro questions for MCQs (left) and Numericals (right) across difficulty levels: $\mathcal{CP}_D$ (Difficult), $\mathcal{CP}_C$ (Challenging), $\mathcal{CP}_M$ (Medium), and $\mathcal{CP}_E$ (Easy) ($\mathcal{CP}_E$). Subfields include Bio, Physical, Organic, and Inorganic chemistry. The majority of questions are concentrated in the $\mathcal{CP}_D$ section.}
    \label{fig:category_dis_supp}
\end{figure*}

\begin{table*}[ht]            
  \centering        
  \caption{\textbf{\textit{Additional Model Performance across Benchmark Sets}}}
  \label{tab:new_models_performance}         
  \begin{tabular}{|l|l cccc cccc|}           
  \toprule          
  & & \multicolumn{4}{c}{\textbf{MCQ Accuracy}} & \multicolumn{4}{c|}{\textbf{Numerical Accuracy (\%)}} \\   
  \cmidrule(lr){3-6} \cmidrule(lr){7-10}  
  Model & Size & $\mathcal{CP}_E$ & $\mathcal{CP}_M$ & $\mathcal{CP}_C$ & $\mathcal{CP}_D$ & $\mathcal{CP}_E$ & $\mathcal{CP}_M$ & $\mathcal{CP}_C$ & $\mathcal{CP}_D$ \\          
  \midrule          
  ChemDFM-v1.5-8B       & 8B  & 0.607 & \textbf{0.519} & 0.405 & \textbf{0.404} & 46.43 & 42.00 & 24.88 &   
  15.34 \\          
  ChemLLM-7B-Chat-DPO   & 7B  & 0.615 & 0.392 & 0.368 & 0.351 & 19.39 & 20.00 & 16.59 & 10.35 \\            
  Qwen3-8B   & 8B  & 0.807 & 0.451 & 0.248 & 0.138 & 45.41 & 25.00 & 22.44 & 10.72 \\            
  \midrule          
  ChemDFM-v2.0-14B      & 14B & \textbf{0.908} & 0.473 & \textbf{0.414} & 0.367 & \textbf{75.00} &          
  \textbf{70.00} & \textbf{48.78} & \textbf{41.27} \\           
  Qwen3-14B  & 14B & 0.867 & 0.464 & 0.301 & 0.201 & 50.00 & 30.00 & 27.32 & 14.21 \\            
  \midrule          
  ChemLLM-20B-Chat-DPO  & 20B & 0.833 & 0.430 & 0.338 & 0.316 & 55.10 & 47.00 & 34.15 & 20.32 \\            
  Qwen3-32B  & 32B & 0.878 & 0.468 & 0.342 & 0.245 & 48.47 & 28.00 & 25.85 & 14.34 \\            
  \bottomrule       
  \end{tabular}     
\end{table*}

\begin{table*}[b]
  \centering
  \caption{Performance on Subfields (Bio-Chemistry and Inorganic-Chemistry)}
  \tiny
  \label{tab:chem_combined1}
  \resizebox{\textwidth}{!}{
    \begin{tabular}{@{}cllll|lllll@{}}
      \toprule
      \multicolumn{5}{c|}{\textbf{Bio-Chemistry}} & \multicolumn{5}{c}{\textbf{Inorganic-Chemistry}} \\
      Category & Size & Top & Avg. & Cnt & Category & Size & Top & Avg. & Cnt \\
      \midrule
      $\mathcal{CP}_E$     & 07B    & 100 & 98 $\pm$ 03 & 9  & $\mathcal{CP}_E$ & 07B    & 92 & 86 $\pm$ 04 & 9 \\
      $\mathcal{CP}_E$     & 10B    & 100 & 100 $\pm$ 00 & 8  & $\mathcal{CP}_E$ & 10B    & 95 & 91 $\pm$ 05 & 8 \\
      $\mathcal{CP}_E$     & 14B    & 100 & 100 $\pm$ 00 & 10 & $\mathcal{CP}_E$ & 14B    & 97 & 91 $\pm$ 05 & 10 \\
      $\mathcal{CP}_E$     & 32B    & 100 & 100 $\pm$ 00 & 9  & $\mathcal{CP}_E$ & 32B    & 96 & 93 $\pm$ 05 & 9 \\
      $\mathcal{CP}_E$     & 70+B   & 100 & 100 $\pm$ 00 & 8  & $\mathcal{CP}_E$ & 70+B   & 97 & 90 $\pm$ 06 & 8 \\
      $\mathcal{CP}_E$     & GPT-4o & 100 & 100 $\pm$ 00 & 3  & $\mathcal{CP}_E$ & GPT-4o & 97 & 91 $\pm$ 08 & 3 \\
      $\mathcal{CP}_M$     & 07B    & 100 & 91 $\pm$ 10 & 9   & $\mathcal{CP}_M$ & 07B    & 92 & 80 $\pm$ 23 & 9 \\
      $\mathcal{CP}_M$     & 10B    & 100 & 100 $\pm$ 00 & 8  & $\mathcal{CP}_M$ & 10B    & 92 & 89 $\pm$ 04 & 8 \\
      $\mathcal{CP}_M$     & 14B    & 100 & 97 $\pm$ 08 & 10  & $\mathcal{CP}_M$ & 14B    & 99 & 78 $\pm$ 29 & 10 \\
      $\mathcal{CP}_M$     & 32B    & 100 & 99 $\pm$ 03 & 9   & $\mathcal{CP}_M$ & 32B    & 92 & 84 $\pm$ 21 & 9 \\
      $\mathcal{CP}_M$     & 70+B   & 100 & 99 $\pm$ 04 & 8   & $\mathcal{CP}_M$ & 70+B   & 93 & 82 $\pm$ 24 & 8 \\
      $\mathcal{CP}_M$     & GPT-4o & 100 & 97 $\pm$ 06 & 3   & $\mathcal{CP}_M$ & GPT-4o & 93 & 88 $\pm$ 07 & 3 \\
      $\mathcal{CP}_C$     & 07B    & 100 & 88 $\pm$ 10 & 9   & $\mathcal{CP}_C$ & 07B    & 66 & 54 $\pm$ 09 & 9 \\
      $\mathcal{CP}_C$     & 10B    & 100 & 91 $\pm$ 08 & 8   & $\mathcal{CP}_C$ & 10B    & 69 & 63 $\pm$ 05 & 8 \\
      $\mathcal{CP}_C$     & 14B    & 100 & 93 $\pm$ 12 & 10  & $\mathcal{CP}_C$ & 14B    & 78 & 65 $\pm$ 12 & 10 \\
      $\mathcal{CP}_C$     & 32B    & 100 & 98 $\pm$ 05 & 9   & $\mathcal{CP}_C$ & 32B    & 71 & 66 $\pm$ 08 & 9 \\
      $\mathcal{CP}_C$     & 70+B   & 100 & 97 $\pm$ 06 & 8   & $\mathcal{CP}_C$ & 70+B   & 76 & 66 $\pm$ 09 & 8 \\
      $\mathcal{CP}_C$     & GPT-4o & 89 & 82 $\pm$ 06 & 3    & $\mathcal{CP}_C$ & GPT-4o & 76 & 66 $\pm$ 16 & 3 \\
      $\mathcal{CP}_D$     & 07B    & 70 & 62 $\pm$ 07 & 9    & $\mathcal{CP}_D$ & 07B    & 55 & 47 $\pm$ 07 & 9 \\
      $\mathcal{CP}_D$     & 10B    & 73 & 68 $\pm$ 03 & 8    & $\mathcal{CP}_D$ & 10B    & 60 & 55 $\pm$ 05 & 8 \\
      $\mathcal{CP}_D$     & 14B    & 84 & 70 $\pm$ 16 & 10   & $\mathcal{CP}_D$ & 14B    & 73 & 59 $\pm$ 14 & 10 \\
      $\mathcal{CP}_D$     & 32B    & 79 & 73 $\pm$ 09 & 9    & $\mathcal{CP}_D$ & 32B    & 67 & 61 $\pm$ 06 & 9 \\
      $\mathcal{CP}_D$     & 70+B   & 81 & 77 $\pm$ 05 & 8    & $\mathcal{CP}_D$ & 70+B   & 68 & 63 $\pm$ 05 & 8 \\
      $\mathcal{CP}_D$     & GPT-4o & 82 & 73 $\pm$ 16 & 3    & $\mathcal{CP}_D$ & GPT-4o & 78 & 61 $\pm$ 22 & 3 \\
      \bottomrule
    \end{tabular}
  }
\end{table*}
\begin{table*}[b]
  \centering
  \caption{Performance on subfields (Organic-Chemistry and Physical-Chemistry)}
  \tiny
  \label{tab:chem_combined2}
  \resizebox{\textwidth}{!}{
    \begin{tabular}{@{}cllll|lllll@{}}
      \toprule
      \multicolumn{5}{c|}{\textbf{Organic-Chemistry}} & \multicolumn{5}{c}{\textbf{Physical-Chemistry}} \\
      Category & Size & Top & Avg. & Cnt & Category & Size & Top & Avg. & Cnt \\
      \midrule
      $\mathcal{CP}_E$  & 07B    & 94 & 86 $\pm$ 05 & 9  & $\mathcal{CP}_E$  & 07B    & 88 & 83 $\pm$ 05 & 9 \\
      $\mathcal{CP}_E$  & 10B    & 96 & 90 $\pm$ 03 & 8  & $\mathcal{CP}_E$  & 10B    & 90 & 87 $\pm$ 04 & 8 \\
      $\mathcal{CP}_E$  & 14B    & 96 & 88 $\pm$ 05 & 10 & $\mathcal{CP}_E$  & 14B    & 95 & 86 $\pm$ 08 & 10 \\
      $\mathcal{CP}_E$  & 32B    & 94 & 91 $\pm$ 05 & 9  & $\mathcal{CP}_E$  & 32B    & 91 & 87 $\pm$ 06 & 9 \\
      $\mathcal{CP}_E$  & 70+B   & 94 & 92 $\pm$ 02 & 8  & $\mathcal{CP}_E$  & 70+B   & 92 & 87 $\pm$ 09 & 8 \\
      $\mathcal{CP}_E$  & GPT-4o & 96 & 87 $\pm$ 10 & 3  & $\mathcal{CP}_E$  & GPT-4o & 92 & 84 $\pm$ 12 & 3 \\
      $\mathcal{CP}_M$  & 07B    & 91 & 81 $\pm$ 18 & 9  & $\mathcal{CP}_M$  & 07B    & 80 & 70 $\pm$ 11 & 9 \\
      $\mathcal{CP}_M$  & 10B    & 94 & 92 $\pm$ 02 & 8  & $\mathcal{CP}_M$  & 10B    & 78 & 76 $\pm$ 02 & 8 \\
      $\mathcal{CP}_M$  & 14B    & 96 & 83 $\pm$ 22 & 10 & $\mathcal{CP}_M$  & 14B    & 98 & 73 $\pm$ 16 & 10 \\
      $\mathcal{CP}_M$  & 32B    & 95 & 88 $\pm$ 14 & 9  & $\mathcal{CP}_M$  & 32B    & 80 & 75 $\pm$ 08 & 9 \\
      $\mathcal{CP}_M$  & 70+B   & 98 & 91 $\pm$ 16 & 8  & $\mathcal{CP}_M$  & 70+B   & 81 & 74 $\pm$ 09 & 8 \\
      $\mathcal{CP}_M$  & GPT-4o & 99 & 94 $\pm$ 05 & 3  & $\mathcal{CP}_M$  & GPT-4o & 79 & 74 $\pm$ 08 & 3 \\
      $\mathcal{CP}_C$  & 07B    & 56 & 51 $\pm$ 04 & 9  & $\mathcal{CP}_C$  & 07B    & 78 & 67 $\pm$ 09 & 9 \\
      $\mathcal{CP}_C$  & 10B    & 60 & 56 $\pm$ 02 & 8  & $\mathcal{CP}_C$  & 10B    & 83 & 77 $\pm$ 05 & 8 \\
      $\mathcal{CP}_C$  & 14B    & 73 & 62 $\pm$ 11 & 10 & $\mathcal{CP}_C$  & 14B    & 83 & 73 $\pm$ 13 & 10 \\
      $\mathcal{CP}_C$  & 32B    & 66 & 62 $\pm$ 05 & 9  & $\mathcal{CP}_C$  & 32B    & 81 & 76 $\pm$ 07 & 9 \\
      $\mathcal{CP}_C$  & 70+B   & 70 & 63 $\pm$ 07 & 8  & $\mathcal{CP}_C$  & 70+B   & 83 & 79 $\pm$ 09 & 8 \\
      $\mathcal{CP}_C$  & GPT-4o & 79 & 66 $\pm$ 18 & 3  & $\mathcal{CP}_C$  & GPT-4o & 81 & 72 $\pm$ 13 & 3 \\
      $\mathcal{CP}_D$  & 07B    & 50 & 42 $\pm$ 06 & 9  & $\mathcal{CP}_D$  & 07B    & 63 & 48 $\pm$ 10 & 9 \\
      $\mathcal{CP}_D$  & 10B    & 54 & 49 $\pm$ 04 & 8  & $\mathcal{CP}_D$  & 10B    & 65 & 56 $\pm$ 08 & 8 \\
      $\mathcal{CP}_D$  & 14B    & 66 & 51 $\pm$ 11 & 10 & $\mathcal{CP}_D$  & 14B    & 74 & 61 $\pm$ 13 & 10 \\
      $\mathcal{CP}_D$  & 32B    & 59 & 53 $\pm$ 06 & 9  & $\mathcal{CP}_D$  & 32B    & 72 & 67 $\pm$ 06 & 9 \\
      $\mathcal{CP}_D$  & 70+B   & 61 & 54 $\pm$ 07 & 8  & $\mathcal{CP}_D$  & 70+B   & 72 & 61 $\pm$ 07 & 8 \\
      $\mathcal{CP}_D$  & GPT-4o & 69 & 55 $\pm$ 17 & 3  & $\mathcal{CP}_D$  & GPT-4o & 78 & 55 $\pm$ 25 & 3 \\
      \bottomrule
    \end{tabular}
  }
\end{table*}

\begin{table*}[ht!]
\centering
\caption{Model Performance on ChemPro Sections, College Chemistry, and High School Chemistry}
\resizebox{\textwidth}{!}{
\begin{tabular}{>{\centering\arraybackslash}p{2.cm}
     >{\centering\arraybackslash}p{2.cm}
     >{\centering\arraybackslash}p{2.cm}
     |>{\centering\arraybackslash}p{2.cm}
     >{\centering\arraybackslash}p{2.cm}
     >{\centering\arraybackslash}p{2.cm}
     >{\centering\arraybackslash}p{2.5cm}}
\toprule
\textbf{Models} & \textbf{High School Chemistry} & \textbf{College Chemistry} & \textbf{$\mathcal{CP}_E$} & \textbf{$\mathcal{CP}_M$} & \textbf{$\mathcal{CP}_c$} $\downarrow$ & \textbf{$\mathcal{CP}_D$} $\downdownarrows$ \\
\midrule
7B         & 95  & 87  & 92  & 94  & 70  & 58 \\
10B        & 98  & 93  & 96  & 97  & 77  & 62 \\
14B        & 98  & 93  & 97  & 98  & 85  & 74 \\
32B        & 99  & 94  & 96  & 99  & 81  & 69 \\
70B        & 99  & 96  & 97  & 99  & 83  & 71 \\
OpenAI-o1  & 100 & 97  & 97  & 99  & 83  & 80 \\
\bottomrule
\end{tabular}
}
\label{tab:model_comparison}
\end{table*}

\end{document}